\documentclass[journal]{IEEEtran}

\usepackage{cite}
\usepackage[ruled,vlined,linesnumbered]{algorithm2e}
\usepackage{algorithmic}
\usepackage{graphicx,times}
\usepackage{amsfonts}
\usepackage{amsmath,amssymb}
\usepackage{mathrsfs}
\usepackage{subfigure}
\usepackage{booktabs}
\usepackage{epstopdf}
\usepackage{colortbl}

\usepackage{multirow,array}

\let\labelindent\relax
\usepackage{enumitem}

\newlength\myindent
\newcommand\bindent{%
	\begingroup
	\setlength{\itemindent}{\myindent}
	\addtolength{\algorithmicindent}{\myindent}
}
\newcommand\eindent{\endgroup}

\begin{document}

\title{Multiform Evolution for High-Dimensional Problems with Low Effective Dimensionality}

\author{Yaqing Hou, Mingyang Sun, Abhishek Gupta,  Yaochu Jin,~\IEEEmembership{IEEE Fellow}, Haiyin Piao, Hongwei Ge, Qiang Zhang 

	\thanks{Yaqing Hou, Mingyang Sun, Hongwei Ge, and Qiang Zhang are with the College of Computer Science and Technology, Dalian University of Technology (DLUT), China 116024 (e-mail: houyq@dlut.edu.cn; mysun@mail.dlut.edu.cn;   hwge@dlut.edu.cn; zhangq@dlut.edu.cn).}
	\thanks{Abhishek Gupta is with the School of Mechanical Sciences, Indian Institute of Technology Goa (IIT Goa), India 403401 (e-mail: abhishekgupta@iitgoa.ac.in).}
	\thanks{Yaochu Jin is with the Chair of Nature Inspired Computing and Engineering, Faculty of Technology, Bielefeld University, 33619 Bielefeld, Germany, and also with the Department of Computer Science, University of Surrey, Guildford, Surrey GU2 7XH, U.K. (e-mail: yaochu.jin@uni-bielefeld.de)}
	\thanks{Haiyin Piao is with the School of electronics and information, Northwestern Polytechnical University, China (e-mail: haiyinpiao@mail.nwpu.edu.cn).}
}

\maketitle
%
%
%

\IEEEpeerreviewmaketitle

\begin{abstract}
In this paper, we scale evolutionary algorithms to high-dimensional optimization problems that deceptively possess a \emph{low effective dimensionality} (certain dimensions do not significantly affect the objective function). To this end, an instantiation of the multiform optimization paradigm is presented, where multiple low-dimensional counterparts of a target high-dimensional task are generated via random embeddings. Since the exact relationship between the auxiliary (low-dimensional) tasks and the target is a priori unknown, a \textit{multiform evolutionary algorithm} is developed for unifying all formulations into a single multi-task setting. The resultant joint optimization enables the target task to efficiently reuse solutions evolved across various low-dimensional searches via \textit{cross-form genetic transfers}, hence speeding up overall convergence characteristics. To validate the overall efficacy of our proposed algorithmic framework, comprehensive experimental studies are carried out on well-known continuous benchmark functions as well as a set of practical problems in the hyper-parameter tuning of machine learning models and deep learning models in classification tasks and Predator-Prey games, respectively.
\end{abstract}

\begin{IEEEkeywords}
	High-dimensional search, evolutionary multi-tasking, transfer optimization, random embeddings.
\end{IEEEkeywords}


\section{Introduction}

Evolutionary algorithms (EAs) \cite{back1996} have been proposed as population-based optimization metaheuristics inspired by Darwin's theory of biological evolution. Owing to their simplicity and generality, a plethora of EAs, e.g., genetic algorithms (GAs) and differential evolution (DE), among others, have been widely applied for solving a variety of optimization problems, including combinatorial, continuous, and dynamic optimization \cite{8356195,8456559}. Despite the widespread potential applicability of EAs, most existing works are found to be restricted to problems of moderate dimensionality \cite{zhan2022survey}. Indeed, it has been reported that the efficacy of an EA gradually reduces as the dimension of the search space increases. Therefore, effectively scaling EAs to high-dimensional optimization problems has recently attracted increased research attention \cite{ma2018survey,liu2021comprehensive,deng2021quantum}.


\textcolor{black}{In past years, dimensionality-reduction techniques have been well-studied in addressing the challenges posed by high-dimensional data in data science. The classical methods such as PCA\cite{Lever2017}, KPCA\cite{10.1007/BFb0020217}, LDA\cite{10.5555/944919.944937}, etc., are mainly based on a large amount of existing data samples to learn the mapping function between high and low dimensions while preserving as much information as possible. However, in EAs, the data samples are generated and updated as the evolution progresses, hence bringing the challenge of data scarcity (in the early stages of evolution) and the added computational cost of updating the mapping functions as more data is collected with every generation.}
Recently, \textit{decomposition} and \textit{embedding} have been reported as two major research directions in scaling EAs to high-dimensional optimization problems \cite{9627116}. On the one hand, decomposition-based methods have been quite successful for separable problems, in which a large dimensional optimization problem can be \emph{split} into a number of manageable low-dimensional sub-problems \cite{li2018cooperative,sun2019hybrid,shi2021coevolutionary}. Among them, the cooperative co-evolution (CC) algorithm optimizes several clusters of interdependent variables in a divide-and-conquer manner, thus effectively solving partially separable problems. Due to the very nature of this technique, the decomposition-based approach may fail to work effectively when the decision variables have complex inter-dependencies. Therefore,  recent research on CC mainly focuses on realizing the automatic decomposition scheme by identifying the interactions between decision variables instead of random or manual decomposition.
However, existing decomposition methods, including global differential grouping (GDG) \cite{2016A}, differential grouping 2 (DG2) \cite{7911173} and efficient recursive differential grouping (ERDG) \cite{9141328} commonly demand additional function evaluations on classifying the variables into suitable groups, which results in the consumption of computational resources.

On the other hand, embedding methods tackle high-dimensional searches under the assumption that certain decision variables in the optimization problems do not affect the objective function significantly \cite{rusu2018meta}. Thus, by \textit{embedding} the high-dimensional search into a bounded low-dimensional problem formulation (see details in Section II), significant savings can be achieved in optimization effort. Examples of this approach include \textit{random projection} \cite{kaban2016toward} and \textit{random embedding} \cite{wang2016bayesian}, which have recently demonstrated notable performance in practical problem-solving. It is recognized that, due to the stochastic nature of the aforementioned methods, independently sampled low-dimensional formulations may lead to varying degrees of congruity with preferred solutions to the underlying (large-scale) target task of interest. For instance, as per the method and assumptions in \cite{wang2016bayesian}, there is a non-zero probability $\gamma$ that the optimum solution of the target task will not be contained within the specified bounds of a particular randomly generated low-dimensional search space. As a result, it is often difficult to reliably ascertain whether a particular random projection/embedding will (or will not) be effective for the optimization problem at hand. Existing studies have attempted to overcome this issue by considering \emph{multiple random embeddings} \cite{qian2016derivative,yang2021parallel}. Specifically, \emph{$N$} different low-dimensional search spaces are independently sampled, and the resulting optimization runs are carried out separately for each case; this has the effect of reducing the probability of failure, i.e., the probability of the target optimum solution not being contained within the bounds of any of the \emph{$N$} independently drawn embeddings, at an exponential rate as $\gamma^{(N)}$. It can, however, be argued that the need for such multiple self-contained optimization runs will tend to be time-consuming - especially if limited computational resources restrict the scope for parallelization - as the search is repeated from scratch each time without leveraging the latent correlations that could exist among distinct embeddings.

Taking this cue, the focus of our research is on devising a more efficient and reliable random embedding method for scaling conventional EAs to high-dimensional optimization problems. Different from existing work, we propose to integrate multiple low-dimensional embeddings into a single (joint) optimization run in the spirit of \emph{evolutionary multitasking} \cite{gupta2016multifactorial,tan2021evolutionary}, such that hidden correlations between alternative formulations may be uncovered and harnessed spontaneously. We highlight that the general idea of concurrently tackling multiple alternate formulations of a single target task, as a way to catalyze overall optimization performance, has only recently been conceived under the label of \emph{multiform optimization} \cite{gupta2018insights}.

Note that, by unifying multiple formulations, the hurdle of identifying the most useful/effective one (which is typically unknown a priori) is automatically by-passed in the multiform optimization paradigm. Most importantly, the salient ability of the multitasking approach to exploiting \emph{mutual} relationships that may be present among optimization tasks - through continuous \emph{omni-directional} transfers of computationally encoded knowledge building blocks \cite{gupta2018insights} - allows one to forgo the tabula rasa scheme of existing procedures; thus, the efficiency and effectiveness of considering multiple random embeddings are enhanced, while retaining a low failure probability of $\gamma^{(N)}$.

The core contributions of this paper can be summarized as follows.

\begin{enumerate}
	\item An evolutionary multitasking-based multiform optimization instantiation is proposed for solving high-dimensional optimization problems with low effective dimensionality. The method is based on generating multiple low-dimensional random embeddings, each of which serves as an alternative problem formulation of the original (high-dimensional) task of interest.
    \item To facilitate the automatic and efficient knowledge transfer among multiple formulations, the realization of a cross-form genetic transfer operator and dynamic computational resource-allocation strategies in multiform optimization is proposed and discussed comprehensively.
	\item To test the efficacy of the proposed method, comprehensive empirical studies are conducted on a set of synthetic continuous optimization functions ($D_{max}=5000$). Two practical applications of the approach are presented for hyper-parameter tuning examples of multi-class support vector machines ($D_{max}=190$) and deep learning models ($D_{max}=5000$) in classification tasks and a classical predator-prey game, respectively.
\end{enumerate}

The rest of this paper is organized as follows: 
Section~\ref{sec:mfo} provides an overview of the basic concepts of random embedding and multiform optimization. In Section~\ref{sec:mfea}, the proposed multiform EAs are presented. To examine the efficacy of the proposed method, comprehensive empirical studies for both synthetic continuous optimization problems as well as practical applications are presented in Sections~\ref{sec:exp} and ~\ref{sec:applications}, respectively. Finally, concluding remarks and plans for future work are presented in Section~\ref{sec:con}.

\section{Basic Concepts}\label{sec:mfo}
In this section, we examine the potential benefits of optimizing a target large-scale black-box function characterized by low effective dimensionality with its multiple alternate formulations. We are particularly interested in addressing the challenges when scaling conventional EAs to such problems.

Let $f:\mathcal{X} \to \mathbb{R}$ be an objective function defined on a region $\mathcal{X} \subseteq \mathbb{R}^D$, where $D$ denotes the dimensionality of the search space $\mathcal{X}$. The problem is assumed to be black-box, implying that there is no closed-form expression available for $f$, and one must query an oracle to obtain the value of $f$ at a specific point $\mathbf{x}$. An optimization problem in the present work is simply stated as
\begin{equation}
	\mathbf{x}^* = \text{argmin}_{\mathbf{x}\in \mathcal{X}} f(\mathbf{x}).
	\label{eq:def_1}
\end{equation}
Without loss of generality,  we consider $\mathcal{X} \subseteq [-1,1]$ (which is always achievable via linear rescaling). 

In what follows, we first discuss the details of the random embedding method; this approach provides a way to generate multiple low-dimensional formulations of a given high-dimensional optimization task of interest. Subsequently, we introduce the concept of the multiform optimization paradigm and present a specific realization in the context of high-dimensional optimization - utilizing the aforementioned random embedding scheme. We demonstrate the benefits of the paradigm through some illustrations based on commonly used synthetic benchmark functions (namely, Ackley and Rastrigin).

\subsection{Random Embedding}
\begin{figure}
\begin{center}
\includegraphics[trim = 0mm 0mm 0mm 0mm,clip, width = 2.6in]{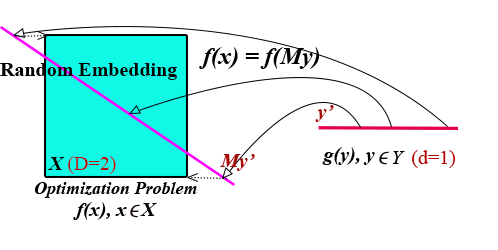}
\caption{Illustration of random embedding from $D=2$ to $d=1$. The box denotes the original constrained solution space $\mathcal{X}$, while the red line illustrates a compressed and bounded low-dimensional search space $\mathcal{Y}$. }
\label{fig:randomembedding}
\end{center}
\end{figure}

As has been noted by many researchers, for certain classes of high-dimensional optimization problems, most dimensions do not affect the objective function significantly. Typical examples include the hyper-parameter optimization of machine-learning models \cite{qian2016scaling,tu2019autone} as well as the automatic configuration of state-of-the-art algorithms for solving NP-hard problems \cite{narayanan2021randomized}. According to \cite{wang2016bayesian}, these optimization problems possess a ``low effective dimensionality.'' Formally, let a function $f:\mathcal{X} \to \mathbb{R}$ be a function defined on a bounded region $\mathcal{X}$, which is considered to have an effective dimensionality $d_e$ (with $d_e<D$), if there exists a linear subspace $\Psi$ of dimensionality $d_e$ such that $\forall \mathbf{x} \in \mathcal{X}$ (see Fig.~\ref{fig:randomembedding}), we have
\begin{equation}
f(\mathbf{x}) = f(\mathbf{x}_e + \mathbf{x}_\bot) = f(\mathbf{x}_e), 
\end{equation}
where $\mathbf{x}_e \in \Psi \subseteq \mathcal{X}$, $\mathbf{x}_\bot \in \Psi^\bot \subseteq \mathcal{X}$, and $\Psi^\bot$ indicates the orthogonal complement of $\Psi$. $\Psi$ is referred to as the \textit{effective subspace} of $f$ and $\Psi^\bot$ denotes the \textit{constant subspace}. 

Equation (2) simply implies that the function $f$ is sensitive only to changes along the effective dimensions $d_e$ in subspace $\Psi$ (i.e., effective subspace). The function value does not change along $\Psi^\bot$ (i.e., constant subspace). In this setting, the method of random embedding \cite{wang2016bayesian} has been shown to be useful for solving high-dimensional optimization problems. Specifically, consider the function $f:\mathcal{X} \to \mathbb{R}$ with effective dimensionality $d_e$ and a random embedding matrix $\mathbf{M}^{D \times d}$ with entries sampled independently from $\mathcal{N}(0,1)$, where $D > d \ge d_e$ and $\mathcal{N}(0,1)$ denotes the standard normal distribution. Then, it can be shown that for any $\mathbf{x} \in \mathcal{X}$ there exists with probability 1 a solution $\mathbf{y}$ such that
\begin{equation}
f(\mathbf{x}) = f(\mathbf{My}).
\end{equation}
A proof of the result in Eq.~(3) is available in \cite{wang2016bayesian}, and is thus not reproduced herein. By extension, for any $\mathbf{x}^* \in \mathcal{X}$, there exists $\mathbf{y}^*$ (unbounded) such that $f(\mathbf{My}^*)= f(\mathbf{x}^*)$. Accordingly, the optimization problem $f(\mathbf{x})$ can be re-formulated into
\begin{equation}
\mathbf{y}^* = \textmd{argmin}_{\mathbf{y}} g(\mathbf{y}),
\label{eq:def_1}
\end{equation}
where $g(\mathbf{y})=f(\mathbf{My})$. As a result, we can optimize the lower-dimensional function $g(\mathbf{y})$ by simply using $\mathbf{M}$ to map the solutions to the original high-dimensional search space.

To make the application of EAs practically viable, a further step of defining bounds for the low-dimensional search space, such that $\mathbf{y} \in \mathcal{Y}$, where the dimensionality of $\mathcal{Y}$ is $d$. This step leads to a slight modification of the formulation in Eq.~(4) as
\begin{equation}
\mathbf{y}^* = \textmd{argmin}_{\mathbf{y}\in \mathcal{Y}} g(\mathbf{y}).
\end{equation}
This is important because the effectiveness of the optimization process depends on the size of $\mathcal{Y}$. If $\mathcal{Y}$ is small, it is easier for the solver to locate the optimum within $\mathcal{Y}$. \textcolor{black}{However, if we configure $\mathcal{Y}$ too small, then there is an increasing probability (labeled as $\gamma$) that the global optimum of the original high-dimensional problem is located outside $\mathcal{Y}$ \cite{wang2016bayesian}}. To handle this issue, a natural approach is to run the optimization algorithm multiple times with different (\emph{independently sampled}) random embedding matrices (as was discussed in the Introduction) \cite{qian2016derivative}.

\subsection{Evolutionary Optimization with Knowledge Transfer} \label{sec:multiform_problem}
To date, most existing EAs tend to start the search process from scratch, with the assumption of a zero prior knowledge state. As a result, in many practical applications, especially those characterized by high-dimensional search spaces, conventional EAs suffer from an exponential growth in the number of solution evaluations needed to start producing high-quality results. To improve the search efficiency, the paradigm of \textit{transfer optimization} has recently been introduced in \cite{gupta2018insights,tan2021evolutionary, 8100935}, which formalized the idea of learning and reusing beneficial information across related optimization tasks. 

Over the past few years, several methods have been proposed for accelerating the search efficiency of EAs by transferring encoded genetic information from previously solved source tasks to a related target task in a sequential manner \cite{tian2019bi,xue2020affine}. More recent studies, such as that in \cite{gupta2016multifactorial}, have introduced the concept of evolutionary multitasking, which aims at tackling distinct optimization tasks (including source and target tasks) concurrently. In this case, the knowledge transfer among different tasks was achieved implicitly through the crossover-based sharing of genetic materials in a unified search space. Nevertheless, the majority of existing methodologies of incorporating automatic knowledge transfer in evolutionary computation are found to deal with \textit{distinct} (or self-contained) optimization tasks \cite{zhong2018multifactorial,liang2021multiobjective}. The transfer of knowledge among multiple alternate formulations of a single target task of interest has been relatively less explored, making it a fertile area for future research investigations. 

\begin{figure}
	\begin{center}
		\includegraphics[trim = 2mm 2mm 2mm 2mm,clip, width = 2.8in]{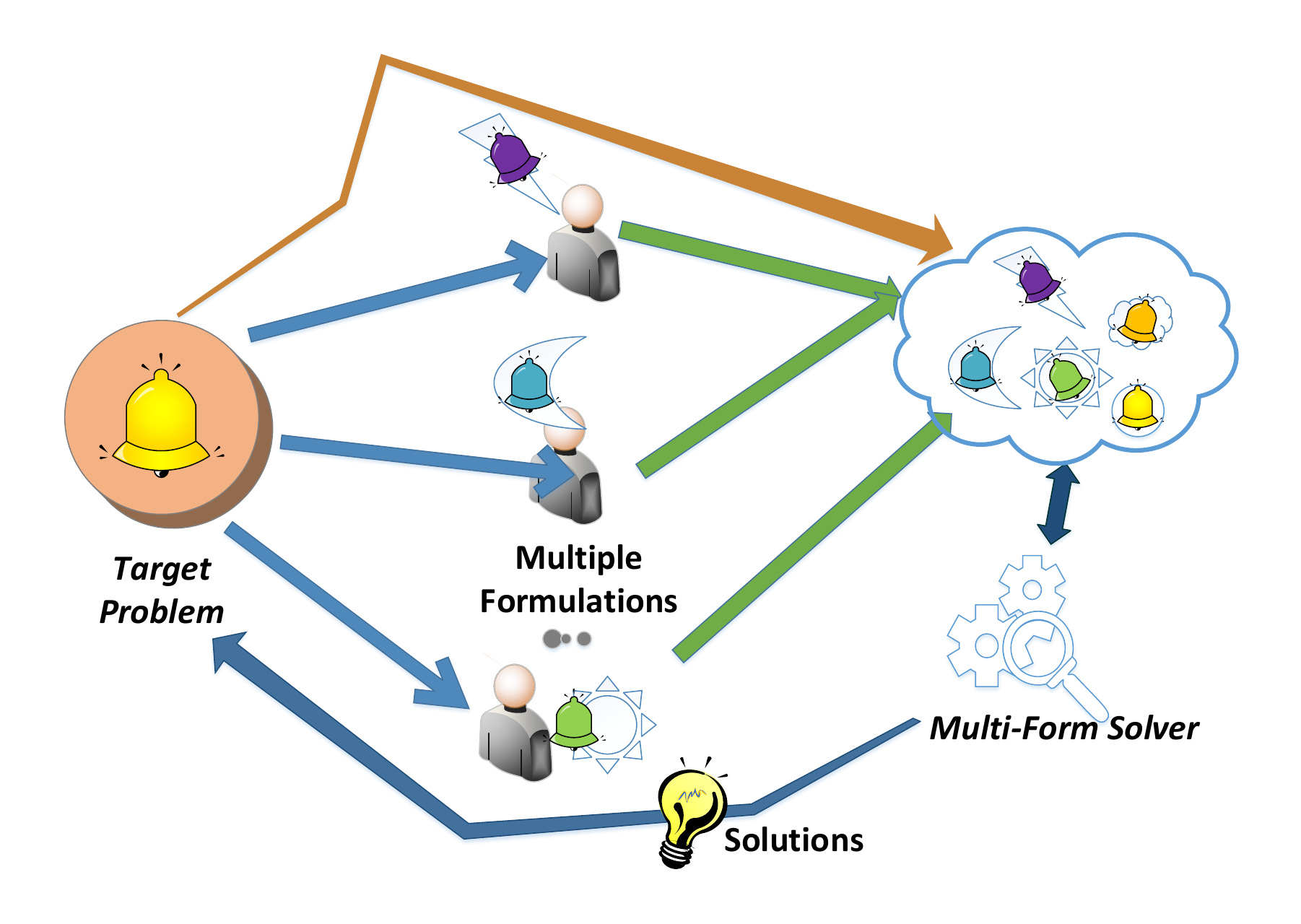}
		\caption{Multiform optimization combines multiple formulations of a target optimization task into one all-encompassing algorithm and optimizes all of the formulations in a concurrent manner.}
		\label{fig:fmfo_structure}
	\end{center}
\end{figure}

Recently, the paradigm of multiform optimization, as a particular type of multitask optimization, has been proposed for solving a target optimization problem by leveraging its alternative formulations (see Fig.~\ref{fig:fmfo_structure} for an illustration of the framework) \cite{gupta2018insights}. To elaborate, a single target optimization problem, labeled $\mathcal{T}$, is assumed to be formulated in different (self-contained) ways; namely, $\Theta_1, \Theta_2,...,\Theta_N$, where $N$ is the number of formulations. We assume that solving any $\Theta_i$ provides an approximate solution to the original problem $\mathcal{T}$, with the quality of the approximation being unknown. Given such a setting, where the aptest formulation (i.e., the one that leads to the best approximation) of the original problem may be unclear to a practitioner, we propose a multitask solver to be deployed to tackle all of the (provided) formulations concurrently, thereby reducing the computational overhead of tackling each formulation from scratch. The overall framework is thus deemed to represent the concept of multiform optimization. Notably, existing studies have proposed to perform search space transformations on the given optimization problem with a dimensionality $D < 1000$, and use the useful information found on the simplified problems to enhance the optimization process in the original problem space \cite{feng2021multi}. Moreover, there are some other recent studies, albeit focusing on specific domains of applications, are also aligned with the idea of multiform optimization \cite{chen2020evolutionary,zhang2021study}.

\subsection{Multiform High-Dimensional Search with Random Embeddings: Problem Specification} \label{sec:multiform_problem}

In the present study, we transform the target optimization formulation (the original task) into a series of low-dimensional formulations (the alternate tasks)  via multiple independent random embeddings. After that, all of the formulations are optimized with the multiform optimization framework. By progressing multiple formulations in tandem, different formulations are able to leverage the beneficial information transferred from one another continuously and automatically. Based on a series of random embedding matrices $\mathbf{M}_1,\mathbf{M}_2,...,\mathbf{M}_N$, the target optimization problem $f(\mathbf{x}), \mathbf{x}\in\mathcal{X}$ is re-formulated into $N$ low-dimensional formulations $\Theta_1, \Theta_2,...,\Theta_N$, belonging to the search space $\mathcal{Y}_1,\mathcal{Y}_2,..,\mathcal{Y}_N$, respectively. $\Theta_n$ is described by an optimization problem of function $g_n$ as
\begin{equation}
\mathbf{y}^*_n = \textmd{argmin}_{\mathbf{y}\in \mathcal{Y}_n} \; g_n(\mathbf{y}).
\label{eq:def_1}
\end{equation}
Since $f(\mathbf{x}) = f(\mathbf{My})$, the high-dimensional optimization problem $f(\mathbf{x})$ thereby can be summarized as
\begin{equation}
\mathbf{y}^* = \textmd{argmin}_{\mathbf{y}_n^*}f(\mathbf{M}_n\mathbf{y}_n^*),n=1,2,...,N.
\label{eq:multiform_problem}
\end{equation}
As discussed in the Introduction, there exists a non-zero probability $\gamma$ that the optimum solution $\mathbf{x}^*$ of the target task will not be located inside the low-dimensional search space $\mathcal{Y}$. In the proposed method, the failure probability when solving the optimization problem in Eq.~\ref{eq:multiform_problem} can be reduced to $\gamma^N$, which converges to $0$ exponentially concerning the number of low-dimensional optimization problems $N$. 

\begin{figure}
	\begin{center}
		\subfigure[Ackley Function ($f(\textbf{x})$).]{\includegraphics[trim = 0mm 0mm 0mm 0mm, clip ,width =1.6 in]{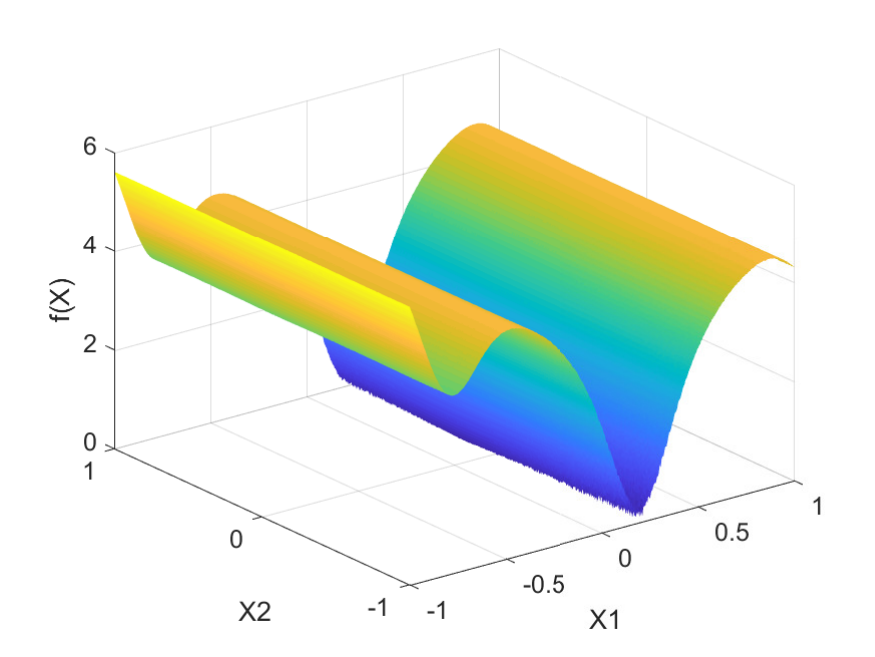}\label{fig:2DAckley}}
		\subfigure[Ackley Functions ($g(\textbf{y})$).]{\includegraphics[trim = 0mm 0mm 0mm 0mm, clip ,width =1.6 in]{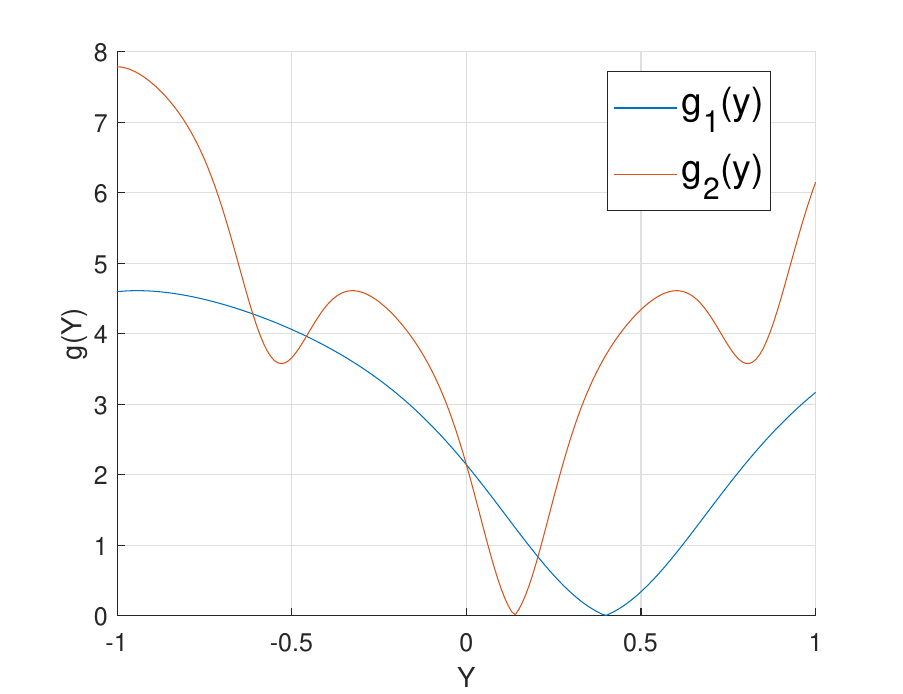}\label{fig:1dAckley}}
		\subfigure[Rastrigin Function ($f(\textbf{x})$).]{\includegraphics[trim = 0mm 0mm 0mm 0mm, clip ,width =1.6 in]{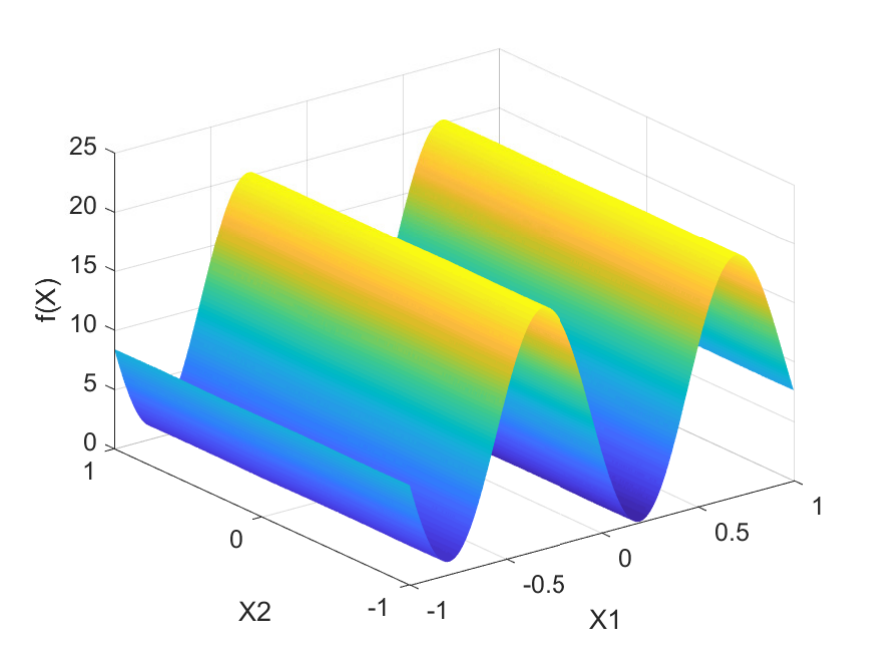}\label{fig:2DRastrigin}}
		\subfigure[Rastrigin Functions ($g(\textbf{y})$).]{\includegraphics[trim = 0mm 0mm 0mm 0mm, clip ,width =1.6 in]{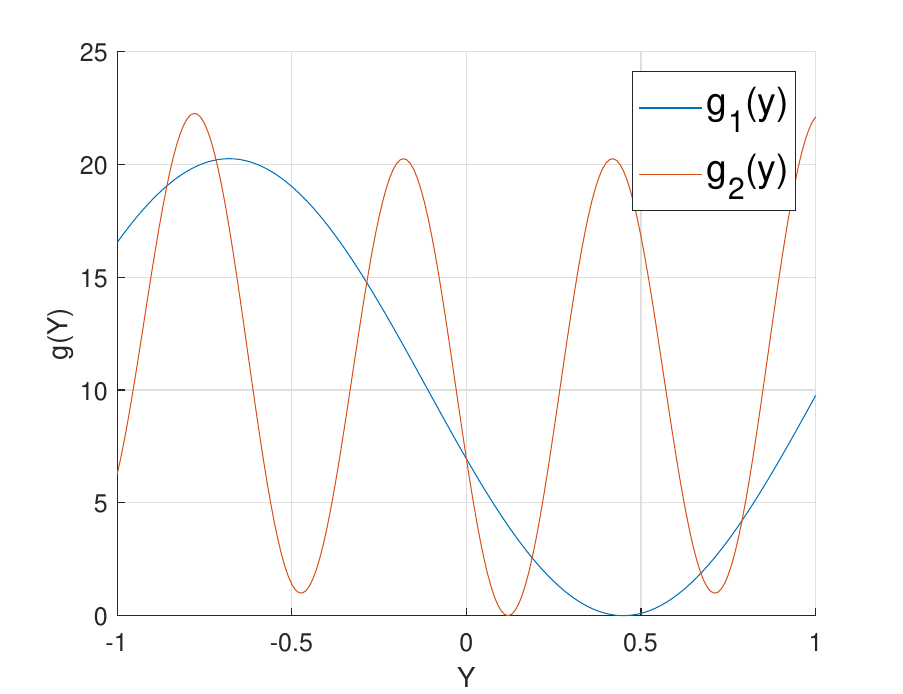}\label{fig:1dRastrigin}}
		
		\caption{Illustrated functions (i.e., Ackley and Rastrigin) in $D=2$ dimensions only have $d=1$ effective dimension. According to random embedding matrices, $f(\textbf{x})$ is formulated into $d$-dimensional functions $g_1(\textbf{y})$ and $g_2(\textbf{y})$, which possess different forms of searching landscapes.}
		\label{fig:FuncDemo}
	\end{center}
\end{figure}

Next, to explore the potential benefits of the proposed multiform high-dimensional optimization with random embeddings, we refer to the illustrative examples of the Ackley and Rastrigin functions. As depicted in Fig.~\ref{fig:FuncDemo}, a function $f(\mathbf{x}), \mathbf{x}\in\mathcal{X}$ in $D=2$ dimensions may have $d=1$ effective dimension; i.e., the objective value does not change with respect to changes in the other dimensions. Given random embedding matrices $\mathbf{M}_1$ and $\mathbf{M}_2$,  $f(\mathbf{x})$ is re-formulated into $g_1(\mathbf{y}), \mathbf{y}\in\mathcal{Y}_1$ and $g_2(\mathbf{y}), \mathbf{y}\in\mathcal{Y}_2$. For simplicity, we set $\mathcal{Y}_1,\mathcal{Y}_2 \in [-1,1]^d$. Notably, $\mathbf{M}_1$ and $\mathbf{M}_2$ are generated randomly and independently, and hence, could induce different search landscape properties in $g_1(\mathbf{y})$ and $g_2(\mathbf{y})$. For example, Fig.~\ref{fig:1dAckley} depicts the search landscape of newly generated formulations $g_1(\mathbf{y})$ and $g_2(\mathbf{y})$ of the Ackley function. Notably, $g_1(\mathbf{y})$ is shown to have a relatively smooth search landscape with a gentle gradient. In contrast, $g_2(\mathbf{y})$ possesses a more rugged landscape with steep gradients. Consequently, optimization algorithms are more likely to get trapped at a local optimum. If $g_1(\mathbf{y})$ and $g_2(\mathbf{y})$ are solved together in the proposed framework, then the continuous transfer of information could enable $g_2(\mathbf{y})$  to leverage the smooth landscape of $g_1(\mathbf{y})$, thereby escaping from local optima. As a result, $g_2(\mathbf{y})$ could converge faster to its global optimum $\mathbf{y}^*$, leading to a more efficient search process of the original optimization problem $f(\mathbf{x}^*)= f(\mathbf{My}^*)$. Notably, the Rastrigin function with one effective dimension also showcases qualitatively similar behavior in Fig.~\ref{fig:FuncDemo}.

\section{Multiform Evolutionary Algorithms}\label{sec:mfea}
So far, we have formulated the target high-dimensional optimization problem into multiple lower-dimensional formulations via multiple random embeddings. Each formulation tends to offer a different yet complementary view of the original optimization problem and hence could serve as a helper task that helps guide the search in the overall multiform framework. Taking this cue, in this section, we further introduce a multiform EA (as a variant of the multitask optimization solver) that combines all of the formulations into an all-encompassing search environment. Herein, each formulation could benefit from others via continuous genetic information transfer. In what follows, we first present the generic structure of the multiform EA. Further, the detailed realization of the cross-form genetic transfer and a simple dynamic resource allocation strategy across alternative formulations are discussed.

\subsection{Framework}
In contrast to multitasking EAs in which different optimization tasks are solved together, the interesting feature of the proposed multiform EA is the focus on effectively solving a single complex (i.e., high-dimensional) optimization task while taking advantage of its alternative formulations.

\begin{figure} 
	\begin{center}
		\includegraphics[trim = 0mm 0mm 0mm 0mm,clip, width = 3.5in]{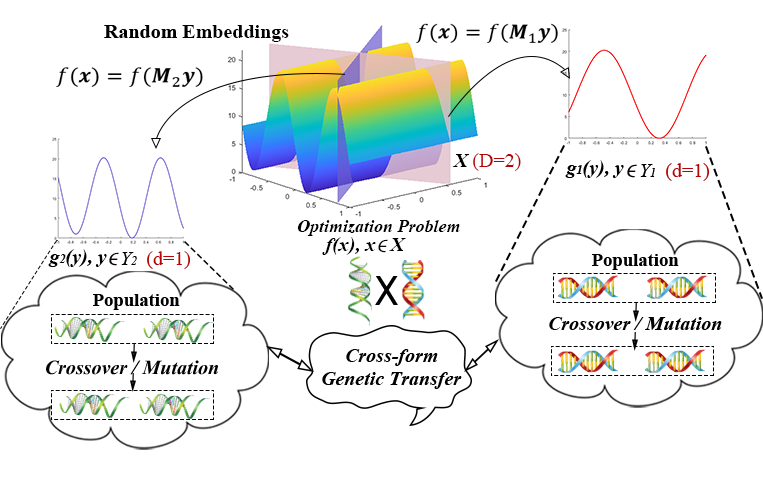}
		\vspace{-0.1 in}
		\caption{Illustration of proposed multiform EA with random embeddings from $D=2$ to $d=1$. Note in this example, the original high-dimensional optimization problem $f(\mathbf{x})$ is formulated into two low-dimensional tasks $g_1(\mathbf{y})$ and $g_2(\mathbf{y})$. The multiform EA enables the cross-form genetic transfer across tasks $g_1(\mathbf{y})$ and $g_2(\mathbf{y})$.}
		\label{fig:MFEA}
	\end{center}
\end{figure}

The pseudo-code of the multiform EA is outlined in Alg.~\ref{alg:framework} and its generic framework is depicted in Fig.~\ref{fig:MFEA}. As illustrated, the multiform EA begins by transforming a high-dimensional optimization problem $f(\mathbf{x}), \mathbf{x} \in \mathcal{X}$ into a series of low-dimensional formulations $ g_1(\mathbf{y}),g_2(\mathbf{y}),...,g_N(\mathbf{y}), \mathbf{y}\in \mathcal{Y}$, which are denoted as $\Theta_1,\Theta_2,...,\Theta_N$. Considering a ``low effective dimensionality'' of $f(\mathbf{x})$, the transformation process is achieved by generating a set of random embedding matrices $\{\mathbf{M}_1,\mathbf{M}_2,...,\mathbf{M}_N\}$, where $\mathbf{M}\in \mathbb{R}^{D \times d}$ and $d$ is the upper bound of the unknown effective dimensionality. Accordingly, the original optimization problem is tackled in a lower-dimensional search space $\mathbf{y}\in \mathcal{Y}$ as $f(\mathbf{M}\mathbf{y}) =f(\mathbf{x})$. Note that there may exist $\mathbf{y}'\in \mathcal{Y}$ such that $\mathbf{M}\mathbf{y}' \notin \mathcal{X}$. If $\mathbf{M}\mathbf{y}'$ is located outside $\mathcal{X}$, then $f(\mathbf{M}\mathbf{y}')$ cannot be evaluated properly. To address this issue, we project $\mathbf{M}\mathbf{y}'$ onto $\mathcal{X}$ by  calculating the shortest Euclidean distance from  $\mathbf{M}\mathbf{y}'$ to the boundary of $\mathcal{X}$ \cite{wang2016bayesian}:
\begin{equation}
\mathbf{x}' = \textmd{argmin}_{\mathbf{x}\in\mathcal{X}}(||\mathbf{x}-\mathbf{M}\mathbf{y}'||_2)).
\label{eq:prjection}
\end{equation}

\begin{algorithm}
	\KwIn{A high-dimensional optimization problem $f(\mathbf{x}), \mathbf{x} \in \mathcal{X}$, the upper bound of the effective dimension $d$ for solving $f(\mathbf{x})$, $d<D$}
	\KwOut{The optimal solution $p$ and the corresponding fitness.}
	\textbf{Begin: Initialization}\\\vspace{1mm}
	\textbf{\enspace Generate} a series of $N$ random embedding matrices $\{\mathbf{M}_1,...,\mathbf{M}_N\}$ where $\mathbf{M}\in \mathbb{R}^{D \times d}$.\\\vspace{1mm}
	\textbf{\enspace Generate} multiform optimization formulations $\Theta_1,\Theta_2,...,\Theta_N$.\\\vspace{1mm}
	\textbf{End}\\\vspace{3mm}
	
	\textbf{Begin: evolutionary evaluation}\\\vspace{1mm}
	\textbf{\enspace Generate} a number of $K$ individuals to form the current population $\mathbf{P}=\{p_1,p_2,...,p_K\}$.\\\vspace{1mm}
	\textbf{\enspace Attribute} individuals in $\mathbf{P}$ to $\{\Theta_1,\Theta_2,...,\Theta_N\}$ respectively $\mathbf{P} = \{p_1^1,p_2^1,...,p_i^2,...,p_K^N\}$.	\\\vspace{1mm}
	\bindent
	\While{stopping criteria are not satisfied}{\vspace{1mm}
		\textbf{Construct} the mappings $\mathbf{W}$ across each pair of formulations (refer to Alg.~\ref{alg:CrossformOptimization}).\\\vspace{1mm}
		\textbf{Perform} variation operators (i.e., crossover and mutation) within each formulation to form the new current population $\mathbf{P}$.\\\vspace{1mm}
		\textbf{Perform} cross-form genetic transfer with each pair of formulations to update $\mathbf{P}$.
	}
	\eindent
	
	\textbf{End}
	\caption{Multiform Evolution}
	\label{alg:framework}
\end{algorithm}

The evolutionary search occurs right after the initialization process. The multiform EA generates a random initial population $\mathbf{P}$ of $K$ individuals in the low-dimensional search space, i.e., $\mathbf{P}=\{p_1,p_2,...,p_K\}$. Each individual in $\mathbf{P}$ is then attributed to a specific formulation randomly. Therefore, we have $\mathbf{P} = \{p_1^1,p_2^1,...,p_i^2,p_{i+1}^2,...,p_K^N\}$, where $N$ is the total number of formulations. In this way, each individual is evaluated with respect to only one formulation among all of the others in the multiform environment. During the search process, the evolutionary optimizer performs variation operators (i.e., crossover and/or mutation) within the sub-populations corresponding to each of the tasks to generate the new current population $\mathbf{P}$. Thereafter, a \textit{cross-form genetic transfer} mechanism is run to update the individual solutions in each sub-population by adapting and transferring the best individuals from the others. To achieve this, the principled connections across each pair of formulations must be constructed beforehand. More detailed discussions on cross-form genetic transfer are presented next.

\subsection{Cross-Form Genetic Transfer}
\textcolor{black}{Multiform optimization} is a specific case of evolutionary multitasking (EMT), where multiple helper tasks of a target task are generated and optimized in an all-encompassing multi-task search environment concurrently. In our work, we utilize the idea of explicit genetic transfer, i.e., EMT via explicit auto-encoding \cite{feng2017autoencoding}\cite{8401802}, to promote the knowledge transfer among tasks (formulations) explicitly using a single-layer denoising autoencoder. Note that when the search landscape or dimensionality of multiple tasks differs greatly, EMT with explicit auto-encoding could enable higher knowledge transfer effectiveness compared to the majority of EMT algorithms that conduct knowledge transfer implicitly through the chromosomal crossover. 

Particularly, in multiform optimization, different formulations (i.e., different random embeddings) give rise to distinct views on the landscape of the original optimization problem; hence, genetic materials from one formulation might lead to poor performance in another formulation \cite{feng2020large}. For example, according to the result in Fig.~\ref{fig:1dRastrigin}, the optimal solution of $g_1(\mathbf{y})$ at approximately $0.4$ leads to poor results if directly transferred to $g_2(\mathbf{y})$. Furthermore, in the case that $\mathcal{Y}_1$ and $\mathcal{Y}_2$ have different dimensionality, the direct exchange of solutions between these two reduced spaces is infeasible. Therefore, it is essential to build a valid connection across such ``\textit{heterologous}'' formulations so that useful patterns in the solutions can be transferred effectively.

To this end, consider two populations of individuals, i.e., $\mathbf{P}^i = \{p_1^i,p_2^i,...,p_Q^i\}$ and $\mathbf{P}^j = \{p_1^j,p_2^j,...,p_Q^j\}$, to represent two different heterologous formulations, say $\Theta_i$ and $\Theta_j$, where $Q$ is the portion of the population size $K$ that is dedicated to formulations $\mathbf{P}^i$ and $\mathbf{P}^j$. Each population is represented as a matrix by stacking all individuals.
Notably, the auto-encoder takes the form of a single layer denoising auto-encoder to build a mapping (or connection) $\mathbf{W}^{ij}$ from task $\Theta_i$ to task $\Theta_j$. $\mathbf{W}^{ij}$ is obtained by minimizing the squared reconstruction loss, which is given by
\begin{equation}
\textmd{argmin}_{\mathbf{W}^{ij}} \sum^Q_{q=1}||\mathbf{W}^{ij}\mathbf{P}^i_q -\mathbf{P}^j_q||^2.
\label{eq:leastsquare}
\end{equation}
$\mathbf{W}^{ij}$ can be solved via the following well-known closed-form expression in Eq.~\ref{eq:mapping}, as follows
\begin{equation}
\mathbf{W}^{ij} = (\mathbf{P}^j{\mathbf{P}^{i}}^\mathsf{T})({\mathbf{P}^i} {\mathbf{P}^i}^\mathsf{T})^{-1},
\label{eq:mapping}
\end{equation}
where ${\mathbf{P}^{i}}^\mathsf{T}$ denotes the matrix transpose of $\mathbf{P}^{i}$. Accordingly, the search information (i.e., the optimized solutions) from $\mathbf{P}^{i}$ can be projected into the population $\mathbf{P}^{j}$ by simply multiplying by $\mathbf{W}^{ij}$, in turn biasing the evolutionary search process of $\mathbf{P}^{j}$. In the case that $\mathbf{P}^{i}$ and $\mathbf{P}^{j}$ have differing dimensionality, $\mathbf{W}$ can be constructed by padding individuals in $\mathbf{P}^{i}$ or $\mathbf{P}^{j}$ with zeros to unify the dimensionality of the solutions \cite{feng2017autoencoding}. Finally, it is worth noting that the genetic transfer based on $\mathbf{W}^{ij}$ is unidirectional, which implies that we must calculate $\mathbf{W}^{ji}$ when transferring solutions from $\mathbf{P}^{j}$ to $\mathbf{P}^{i}$. For a clearer understanding, detailed steps of the cross-form genetic transfer are provided in Alg.~\ref{alg:CrossformOptimization}.

\begin{algorithm}
	\KwIn{$N$ formulations of low-dimensional optimization problems ${\Theta_1,\Theta_2,...,\Theta_N}$}
	\KwOut{The populations $\mathbf{P}^i$ and $\mathbf{P}^j$ after genetic transfer}
	\textbf{Begin: evolutionary search}\\\vspace{1mm}
	\textbf{\enspace Consider} populations $\mathbf{P}^i$, $\mathbf{P}^j$ from $\Theta_i$, $\Theta_j$ ($\Theta_i \ne\Theta_j$).\\\vspace{1mm}
	/*Mapping Construction*/ \\\vspace{1mm}
	\textbf{\enspace Construct} the mappings $\mathbf{W}^{ij}$ and $\mathbf{W}^{ji}$ across populations $\mathbf{P}^i$ and $\mathbf{P}^j$ via Eq.\ref{eq:mapping}.\\\vspace{1mm}
	\textbf{\enspace Choose} the best optimized individuals $p_{best}^i$ and $p_{best}^j$ from $\mathbf{P}_i$ and $\mathbf{P}_j$.\\\vspace{1mm}
	\textbf{\enspace Calculate} the transferable optimized individuals $\widehat{p_i} = \mathbf{W}^{ji}p_{best}^j$ and $\widehat{p_j} = \mathbf{W}^{ij} p_{best}^i$.\\\vspace{1mm}
	\textbf{\enspace Select} individuals $\widetilde{p_i}$ and $\widetilde{p_j}$ of lowest fitness values from $\mathbf{P}_i$ and $\mathbf{P}_j$.\\\vspace{1mm}
	\textbf{\enspace Replace} $\widetilde{p_i} \in \mathbf{P}^i$ and $\widetilde{p_j}\in \mathbf{P}^j$ by $\widehat{p_i}$ and $\widehat{p_j}$.\\\vspace{1mm}
	\textbf{\enspace Continue} the evolution search (i.e., crossover and mutation) with current populations $\mathbf{P}^i$ and $\mathbf{P}^j$.\\
	\textbf{End}
	\caption{Cross-form Genetic Transfer}
	\label{alg:CrossformOptimization}
\end{algorithm}

As outlined, considering populations $\mathbf{P}^i$ and $\mathbf{P}^j$ from two formulations $\Theta_i$ and $\Theta_j$, where $i\ne j$, the cross-form genetic transfer process first calculates the connections $\mathbf{W}^{ij}$ and $\mathbf{W}^{ji}$ across populations $\mathbf{P}^i$ and $\mathbf{P}^j$. Then, for simplicity, we select the best optimized individuals $p_{best}^i$ and $p_{best}^j$ from $\mathbf{P}_i$ and $\mathbf{P}_j$ and map them into the solution spaces of one another via $\widehat{p_i} = \mathbf{W}^{ji}p_{best}^j$ and $\widehat{p_j} = \mathbf{W}^{ij} p_{best}^i$. The optimized solutions $\widehat{p_i}$ and $\widehat{p_j}$ are then merged into their corresponding population space. We facilitate the transfer by replacing the individuals $\widetilde{p_i}$ and $\widetilde{p_j}$ of the lowest fitness values by the knowledge-induced solutions $\widehat{p_i}$  and  $\widehat{p_j}$.  Once this is completed, the evolutionary search process continues with the natural selection step. During the entire process, if a transferred individual is useful for the recipient task, then it survives in the population and goes on to produce similarly useful offspring in future generations. However, if a transferred individual is not beneficial for the recipient task, then it is automatically ejected from the population as a consequence of the evolutionary selection pressure; as a result, the threat of any harmful (or negative) transfers is naturally kept under control.

\textcolor{black}{
It is worth noting that our algorithm involves additional computations in two parts: Cross-form Genetic Transfer and the reverse process of random encoding, but their cost is considered to be negligible compared to Fitness Evaluation. During the genetic transfer process, the derived denoising autoencoder has a closed-form solution, resulting in minimal computational overhead in the evolutionary search process \cite{feng2017autoencoding}. Random encoding is simply a multiplication between two matrices of $N \times d$ and $d \times D$.
}

\subsection{Dynamic Resource Allocation}\label{sec:resource_allocation}
Notably, different alternate formulations tend to offer different views of the target problem, and they usually have different degrees of difficulty~\cite{gong2019evolutionary}. 
\textcolor{black}{To effectively distribute limited computing resources among varied tasks, there have been some studies on computational resource allocation in multitask optimization. One example is based on a normalized attainment function and a multi-step nonlinear regression for evaluating the convergence status of multi-objective optimization problems \cite{9570733}. Probabilistic search distribution modeling has also been studied to inform the allocation of computational resources \cite{9493747,9950429,9943989}}.

In what follows, we propose an efficient and lightweight method for dynamic resource allocation across different formulations during the evolutionary multiform optimization process.

First, a preference function $H_t(k)$ is devised as guidance for estimating the convergence trend of the $k^{th}$ formulation at the $t^{th}$ population generation. Then, the proportion of resources occupied by each formulation can be determined by the following formula:
\begin{equation}
P_k = \frac{e^{H_t(k)}}{\sum_{n \in N}e^{H_t(n)}},
\end{equation}
where $N$ is the number of all formulations in multiform optimization. Notably, the preferences of formulations are mapped into Boltzmann’s acceptance probabilities by a softmax function. The formulation with a higher value of task preference will be assigned more computing resources. 

Note that the objective of dynamic resource allocation is to find a more appropriate $P_k$ for the $k^{th}$ formulation during the optimization process. Taking this cue, we then introduce a simple greedy strategy to adaptively adjust the distribution of $P$ based on the convergence trend of each formulation in regard to  their fitness improvement, which is formulated as follows.
\begin{equation}
C_k = \frac{|f_k(x^{*}(t-1))-f_k(x^{*}(t))|}{f_k(x^{*}(t))+ \epsilon},
\end{equation}
where $f_k(x^*(t))$ is the best fitness value obtained in the $t^{th}$ generation and $\epsilon$ is a small positive number. By taking into consideration the fitness improvement, the formulations with a lower degree of difficulty, which usually converge to a global (local) optimum more quickly, tend to have a lower fitness improvement as the search process progresses, and hence, are expected to be allocated with less computing resources.

\textcolor{black}{$H_t(k)$ does not need to be specially defined for each formula and $H_0(k)$ is initialized to 0.} After each generation, the preference function of each task will be updated as below:
\begin{equation}
H_{t+1}(k) = H_{t}(k) + \alpha(C_k - P_k(\sum_{n \in N}C_n)),
\end{equation}
where $\alpha>0$ indicates the step size for updating $H_t(k)$. In the present study, we set $\alpha=2$, after conducting a simple grid search in $(0, 10]$. Moreover, $P_k(\sum_{n \in N}C_n)$ is the weighted average of the convergence trends of the formulation $k$, which serves as a benchmark item for comparison against $C_k$. If $C_k > P_k(\sum_{n \in N}C_n)$, the preference value of the formulation will increase, and more computing resources will be assigned accordingly. 

\section{Empirical Study}\label{sec:exp}
To verify the efficacy of the evolutionary multiform optimization paradigm for solving high-dimensional problems with low effective dimensionality, comprehensive experimental studies are conducted in this section. In what follows, the performance of the proposed multiform EA is tested on a set of synthetic continuous optimization functions.

\begin{table*}[htbp]
  \centering
  \caption{Benchmark Functions}
    \begin{tabular}{llcc}
    \toprule
    \toprule
          & Name  & Search Range & \multicolumn{1}{c}{Function} \\
    F1    & Ackley & $[-32, 32]$ & $-20e^{(-0.2\sqrt{\frac{1}{D}\sum^D_{i=1}x^2_i})}-e^{\frac{1}{D}\sum^D_{i=1}\cos(2\pi x_i)} + 20 + e$  \\ 
    F2    & Rastrigin & $[-5, 5]$ & \textcolor{black}{$10D + \sum_{i=1}^D\left( x_i^2-10 cos(2\pi x_i)\right)$}  \\
    F3    & Weierstrass & $[-0.5, 0.5]$ & $\sum^D_{i=1}(\sum^{20}_{k=0}0.5^k \cos(2\pi 3^k(x_i+0.5))) - D\sum^{20}_{k=0}0.5^k \cos(2\pi 3^k(0.5))$\\
    F4    & Rosenbrock & $[-5, 5]$ & \textcolor{black}{$\sum_{i=1}^{D-1} \left( 100 ( x_{i+1}-x_i^2)^2 + ( 1-x_i)^2 \right)$} \\
    F5    & Griewank & $[-500, 500]$ & $\sum^D_{i=1}\frac{x^2_i}{4000} - \prod^{D}_{i=1}\cos(\frac{x_i}{\sqrt{i}})+1$ \\
    F6    & Elliptic & $[-5, 5]$ & $\sum^D_{i=1}(10^{6(i-1)/(D-1)} x_i^2)$ \\
    \bottomrule
    \bottomrule
    \end{tabular}%
  \label{tab:benchmark}%
\end{table*}%

\subsection{Experimental Configuration}\label{sec:exp_A}
\textcolor{black}{We begin with choosing some popular continuous optimization functions as shown in Table~\ref{tab:benchmark}\footnote{We release the relevant codes at: https://github.com/Sunrisulfr/Multiform.git}}.
\textcolor{black}{Specifically, all functions possess an effective dimensionality $d_e =30$, embedded in a $D$-dimensional space with $D=5000$. That is, the remaining $D-d_e$ dimensions negligibly affect the values of these functions. For the fairness of the experimental comparison and to keep consistent with the settings of the classical CEC Benchmarks, all embedded functions are rotated based on randomly generated rotation matrices of size $D \times D$ for each run. The global optimum of these functions is randomly generated as well, such that the optimum is shifted from zero.}
In this study, random embedding is employed to reformulate the high-dimensional optimization function into multiple low-dimensional formulations, in which the search scope $\mathcal{Y}$ of each formulation is $[-1,1]^d$ for simplicity and $d$ is assumed to be the upper bound of the effective dimensionality $d_e$. As $d_e$ is a priori unknown, $d$ can be set as different values for different formulations.  \textcolor{black}{For each formulation, only $d$ dimensions are updated during the search process, but the fitness functions are evaluated in all $D$ dimensions by mapping each individual to the solution of the original problem with the embedding matrix $\mathbf{M}$.}

With the aim of scaling conventional EAs to solving high-dimensional optimization problems, in the present study, we employ a classical \textit{DE} \cite{storn1997} as our baseline optimization solver. Without loss of generality, other basic solvers, \textcolor{black}{ i.e., \textit{GA}\cite{Deb1995SimulatedBC}}, can also be applied seamlessly. Denoting \textit{S} as the basic evolutionary solver, four variants of \textit{S} are studied in the experiments.
\begin{enumerate}
	\item The optimization solver \textit{S} in which no dimensionality-reduction techniques are involved. In other words, the solver \textit{S} optimizes the objective function directly in the high-dimensional search space.
	\item The second is a solver \textit{S} with a series of low-dimensional formulations generated by multiple independent random embeddings \cite{qian2016derivative} (labeled as \textit{S+M}), i.e., no genetic transfer and dynamic resource allocation strategies across formulations is carried out. 
	\item The third is a solver \textit{S} with the proposed multiform optimization scheme; i.e., the high-dimensional problem is reformulated into a series of low-dimensional problems. The original function and generated formulations are optimized together with the proposed cross-form genetic transfer method (labeled as \textit{S+MT}), i.e., no dynamic resource allocation strategies are introduced. 
	\item Finally, we consider a solver \textit{S} equipped with the multiform optimization scheme, including both the proposed cross-form genetic transfer and dynamic resource allocation strategies (labeled as \textit{S+MF}).
\end{enumerate}
For a fair comparison, the configuration of evolutionary operators and parameters in \textit{S}, \textit{S+M}, \textit{S+MT}, and \textit{S+MF} are kept the same. Further, the specific experimental settings with respect to the evolutionary solvers are outlined as follows.
\begin{enumerate}
	\item Population size $K=100$,
	\item Maximum function evaluations $Max(FEs)=50,000$,
	\item Independent experimental runs $runs=20$,
	\item Parameter settings in \textcolor{black}{ \textit{DE}}: $CR=0.9$, $F=0.35$.
\end{enumerate}
All algorithms are estimated based on fitness values of function evaluation. In particular, to obtain statistically sound conclusions, the Wilcoxon Signed-Rank test with a $95\%$ confidence interval is conducted on experimental results for comparing and analyzing the performance of different optimization methods.

\subsection{Performance Comparison against Existing Methods}
First, we investigate the performance of our proposed multiform EA for solving high-dimensional optimization problems when compared to the existing methods, including two state-of-the-art evolution strategies, namely \textit{LM-CMA}  and \textit{OpenAI-ES}, a competitive swarm optimizer \textit{CSO}, two popular and novel cooperative co-evolution methods \textit{DECC-G} and \textit{CMA-ES-ERDG}, for real-valued optimization. \textcolor{black}{We perform a grid search on two key parameters of \textit{DE}, i.e., scaling factor $F\in [0,2]$ and crossover rate $CR \in [0,1]$, respectively. }
The learning rate $\alpha$ of \textit{OpenAI-ES} is set to 0.1 after searching on $(0, 0.5]$. For \textit{DECC-G}, \textit{CSO}, \textit{LM-CMA}, and \textit{Seq-CMA}, as they are specially designed for large-scale optimization problems, we choose the hyper-parameter values as recommended in their corresponding papers or released codes.
In addition, the cost of FEs used for decomposing and grouping decision variables in \textit{ERDG} is not accounted for in the optimization. The complete results in terms of objective values across $20$ independent runs with $50,000$ function evaluations are tabulated in Table~II. Notably, symbols  ``$+$'', ``$\approx$'," and ``$-$'' denote that certain algorithms (i.e., \textit{DE}, \textit{DECC-G}, \textit{LM-CMA}, \textit{CSO}, \textit{OpenAI-ES}, \textit{CMA-ES-ERDG}) are statistically better than, similar to and worse than the proposed method (i.e., \textit{DE+MF}). 
The multiform optimization methods in the present experiment embrace $5$ formulations (including four low-dimensional ones generated via random embeddings and one original optimization problem). For simplicity, in this set of experiments, the target problem is configured to possess an effective dimensionality $d_e= 30$, and all low-dimensional formulations bear a common upper bound of the effective dimensionality as $d = 50$.

\newsavebox{\multiformsamD}
\begin{lrbox}{\multiformsamD}
	\renewcommand{\arraystretch}{1.1}
  \begin{tabular}{cc|cccc|ccccc}
    \toprule
    \toprule
    \multicolumn{2}{c}{\multirow{2}[1]{*}{Func}} & \multirow{2}[1]{*}{DE} & \multirow{2}[1]{*}{DE+M} & \multirow{2}[1]{*}{DE+MT} & \multicolumn{1}{c}{\multirow{2}[1]{*}{DE+MF}} & \multirow{2}[1]{*}{DECC-G} & \multirow{2}[1]{*}{CSO} & \multirow{2}[1]{*}{OpenAI-ES} & \multirow{2}[1]{*}{CMA-ES-ERDG} & \multirow{2}[1]{*}{LM-CMA} \\
    \multicolumn{2}{c}{} &       &       &       & \multicolumn{1}{c}{} &       &       &       &   & \\
    \midrule
    \multicolumn{1}{c}{\multirow{2}[0]{0.35cm}{F1}} & Mean   & 4.00E+00 ($-$) & 1.18E+01 ($-$) & 4.30E+00 ($-$) & \textbf{3.03E-02} & 4.68E+00 ($-$) & 6.91E+00 ($-$) & 1.69E+00 ($-$) & 6.39E+00($-$) & 2.53E+00 ($-$)\\
          & Std.   & 8.88E-01 & 1.43E+00 & 1.12E+00 & 7.38E-03 & 7.63E-01 & 3.57E-01 & 9.41E-02 & 1.75E-01 & 1.02E-01\\
    \midrule
    \multicolumn{1}{c}{\multirow{2}[0]{0.35cm}{F2}} & Mean   & 1.97E+02 ($-$) & 9.61E+01 ($-$) & 5.40E+01 ($-$) & 3.50E+01 & 2.61E+02 ($-$)  & 1.65E+02 ($-$) & 1.44E+02 ($-$) & 1.76E+02($-$)& \textbf{3.40E+01} ($\approx$) \\
          & Std.   & 9.79E+00 & 1.96E+01 & 2.18E+01 & 9.40E+00 & 3.48E+01 & 9.53E+00 & 8.52E+00 & 8.39E+00& 5.12E+00 \\
    \midrule
    \multicolumn{1}{c}{\multirow{2}[0]{0.35cm}{F3}} & Mean  & 1.24E+01($-$) & 1.71E+01($-$) & 1.10E+01($\approx$) & \textbf{7.52E+00} & 1.77E+01($-$)  & 1.43E+01
($-$) & 4.03E+01($-$) & 1.33E+01($-$)& 1.87E+01($-$) \\
          & Std.  & 2.82E+00 & 1.48E+00 & 4.69E+00 & 3.16E+00 & 1.48E+00 & 4.89E-01 & 8.50E-01 & 1.28E-01& 1.99E+00\\
    \midrule
    \multicolumn{1}{c}{\multirow{2}[0]{0.35cm}{F4}} & Mean  & 5.14E+01($-$) & 2.97E+03($-$) & 1.23E+02($-$) & \textbf{2.77E+01} & 1.32E+02($-$)  & 2.20E+02($-$) & 1.24E+02($-$) & 1.85E+02($-$) & 2.04E+02($-$) 
\\
          & Std.   & 2.80E+01 & 1.26E+02 & 9.83E+01 & 3.58E+00 & 2.00E+01 & 2.97E+01 & 7.91E+00 & 1.66E+01 & 1.83E+01\\
    \midrule
    \multicolumn{1}{c}{\multirow{2}[0]{0.35cm}{F5}} & Mean   & 1.31E+00($-$) & 3.99E+01($-$) & 1.14E+00($-$) & \textbf{1.68E-02} & 3.23E+00($-$)  & 5.58E+00($-$) &  4.96E-02($-$) & 4.78E+00($-$) & 2.10E-02($\approx$) 
 \\
          & Std.  & 2.35E-01 & 2.49E+00 & 2.46E-01 & 1.31E-02 & 5.26E-01 & 5.75E-01 & 4.94E-03 & 3.30E-01 & 1.21E-02 \\
    \midrule
    \multicolumn{1}{c}{\multirow{2}[1]{0.35cm}{F6}} & Mean   & 1.29E+05($-$) & 8.94E+04($-$) & 1.23E+04($-$) & \textbf{4.05E+03} & 6.11E+04($-$) & 6.64E+04($-$) &  1.54E+04($-$) & 5.62E+04($-$) & 3.79E+04($-$) \\
          & Std.   & 6.05E+04 & 1.43E+04 & 3.09E+03 & 1.85E+03 & 1.85E+04 & 9.12E+03 & 3.44E+03 & 3.31E+03 & 6.79E+03\\

    \bottomrule
    \bottomrule
    \end{tabular}%
    
\end{lrbox}

\begin{table*}[!htb]
	\centering  \caption{Objective values obtained by the multiform EA (\textit{DE+MF}) for solving optimization functions when compared to other methods across $20$ independent runs with $50,000$ function evaluations (``$+$'',``$\approx$,'' and ``$-$'' denote that the corresponding algorithm is statistically significantly better than, similar to, and worse than \textit{DE+MF}, respectively.)} \label{tab_multiformsamD}\scalebox{0.8}{\usebox{\multiformsamD}}
\end{table*}

\begin{figure*}[!htb]
	\begin{center}
		\subfigure[Ackley Function ($D = 5000$)]{\includegraphics[trim = 25mm 90mm 30mm 90mm,  clip,width=0.27\linewidth ]{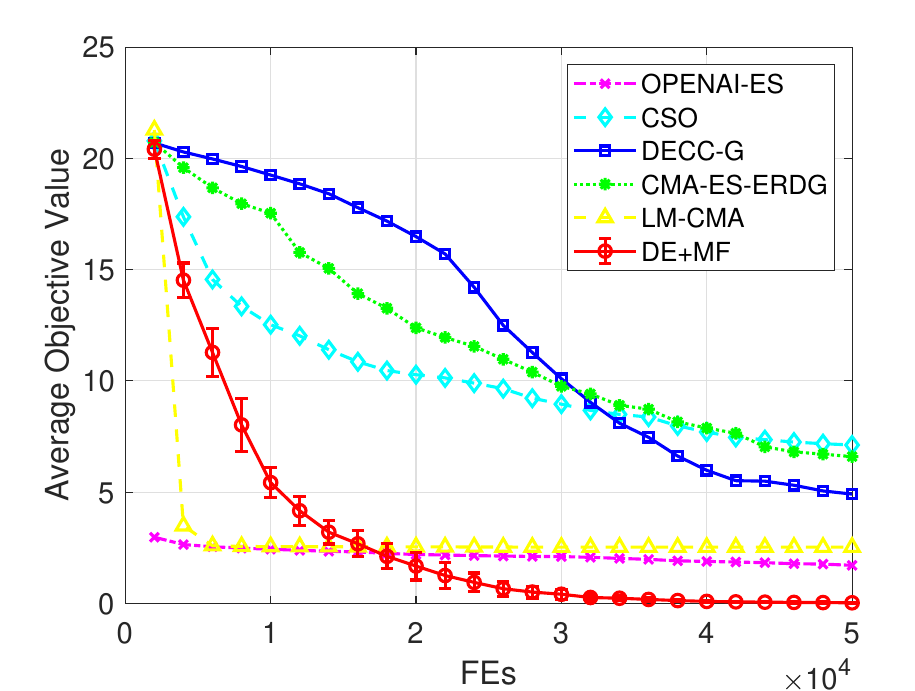}}
		\subfigure[Rastrigin Function ($D = 5000$)]{\includegraphics[trim = 25mm 90mm 30mm 90mm,  clip,width=0.27\linewidth ]{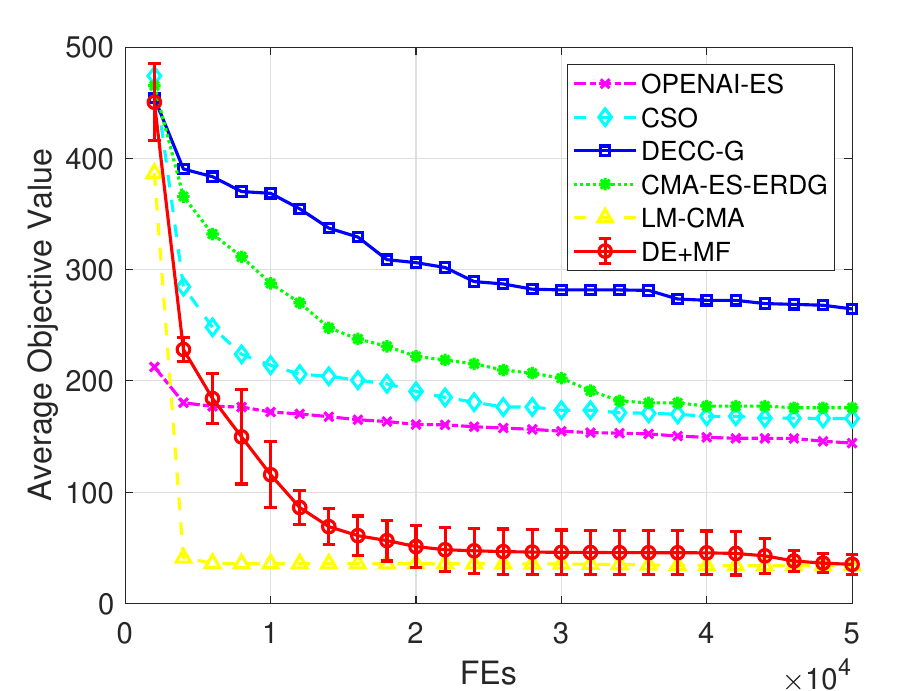}}
		\subfigure[Weierstrass Function ($D = 5000$)]{\includegraphics[trim = 25mm 90mm 30mm 90mm,  clip,width=0.27\linewidth ]{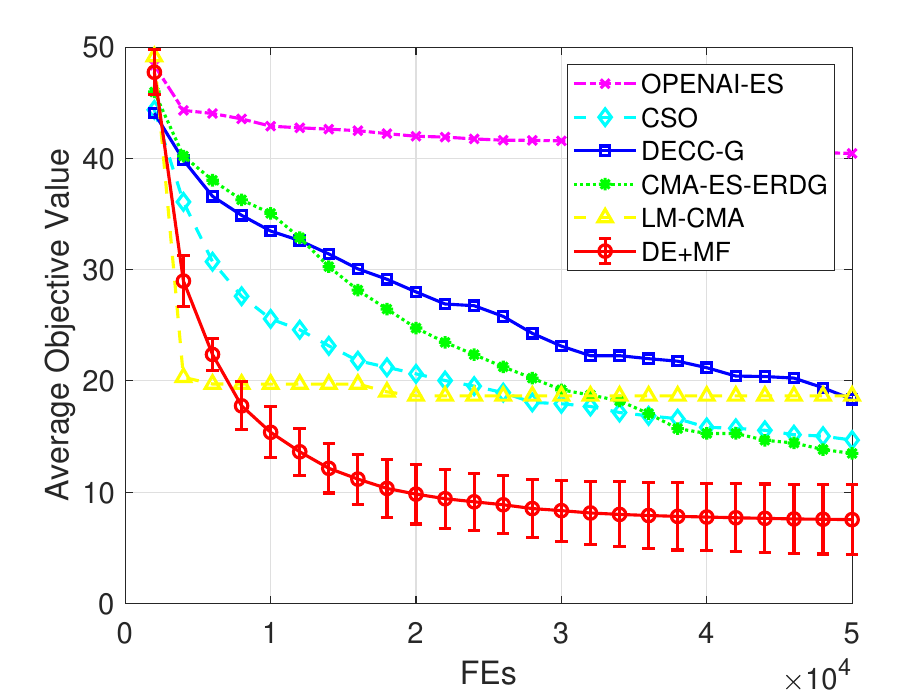}}
		\subfigure[Rosenbrock Function ($D = 5000$)]{\includegraphics[trim = 25mm 90mm 30mm 90mm,  clip,width=0.27\linewidth ]{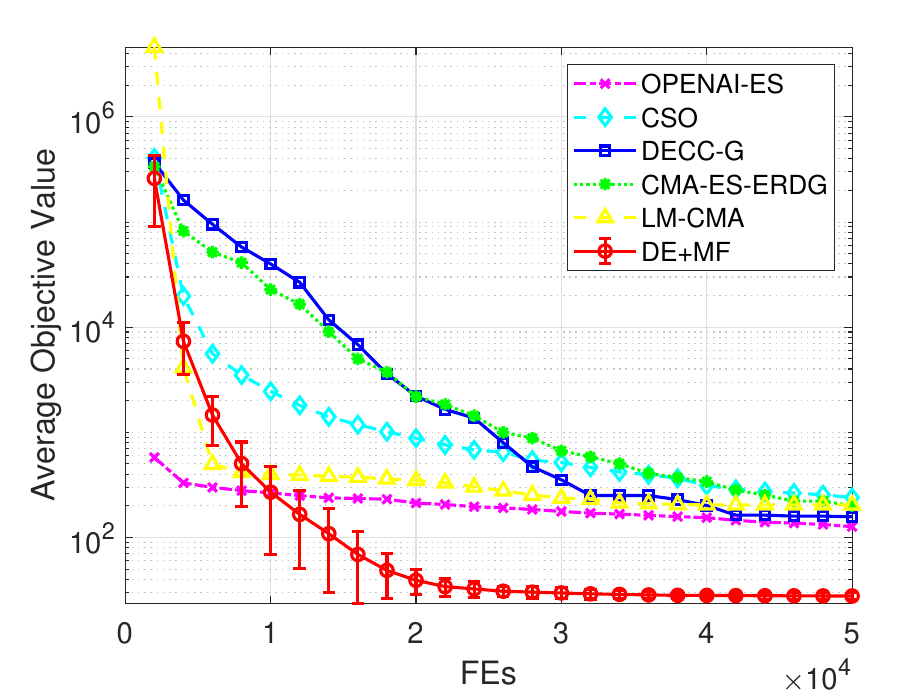}}
		\subfigure[Griewank Function ($D = 5000$)]{\includegraphics[trim = 25mm 90mm 30mm 90mm,  clip,width=0.27\linewidth ]{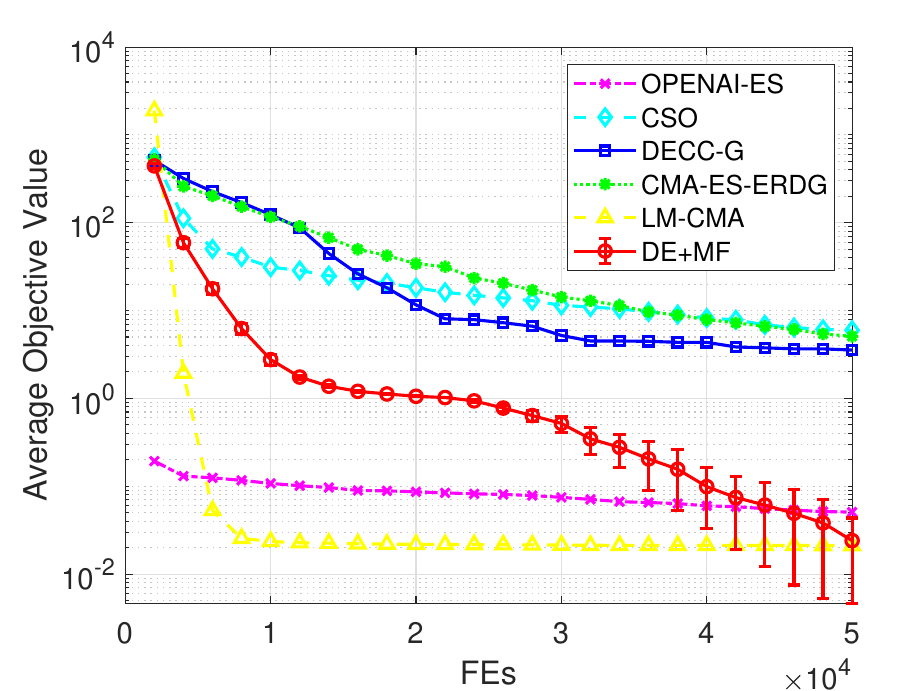}}
		\subfigure[Elliptic Function ($D = 5000$)]{\includegraphics[trim = 25mm 90mm 30mm 90mm,  clip,width=0.27\linewidth ]{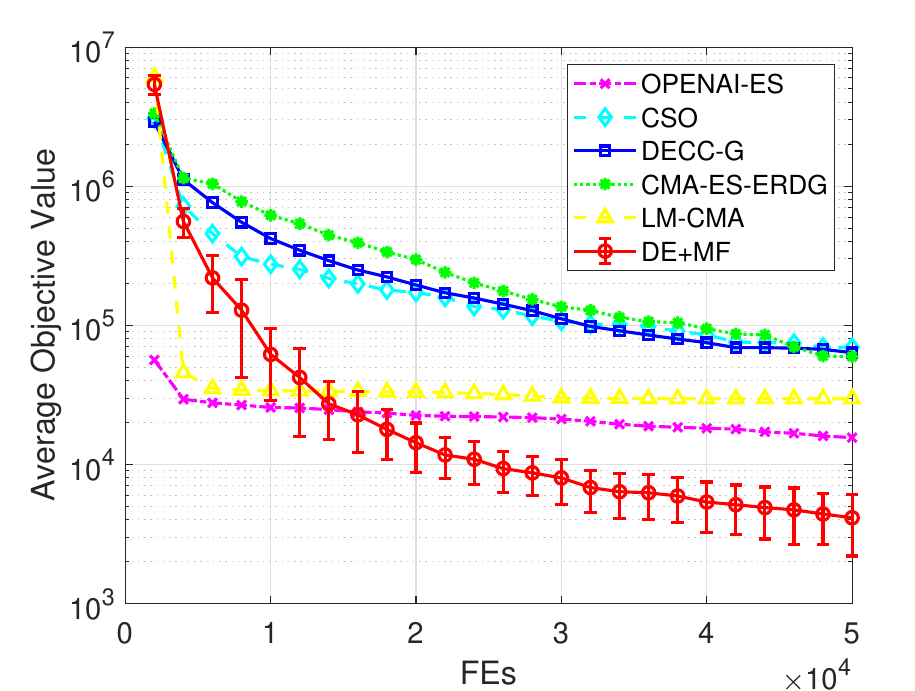}}
		\caption{Objective values obtained by \textit{DECC-G},  \textit{LM-CMA}, \textit{CSO}, \textit{OpenAI-ES}, \textit{CMA-ES-ERDG} and proposed multiform EAs on solving optimization functions averaged by 20 independent runs.}
		\label{fig:comparisons}
	\end{center}
\end{figure*}

According to the numerical results in Table~II, we can see that the proposed multiform EA (i.e., \textit{DE+MF}) obtained significantly better performance in terms of objective values than existing methods, i.e., \textit{DECC-G}, \textit{CMA-ES-ERDG}, \textit{LM-CMA}, \textit{CSO} and \textit{OpenAI-ES}, on six optimization tasks after $50,000$ FEs. These results indicate that with the increasing dimensionality of the optimization tasks (i.e., $D=5000$), it becomes difficult for cooperative co-evolution methods (i.e., \textit{DECC-G} and \textit{CMA-ES-ERDG}) to capture the dependence among variables and decompose them into manageable sub-components. 
Moreover, the evolution strategy-based method (i.e., \textit{OpenAI-ES}) usually has the disadvantages of high complexity and premature stagnation. Even \textit{LM-CMA}, which is designed for large-scale continuous optimization, can only achieve similar optimization performance as \textit{DE+MF} on two test functions while reporting deteriorated performance on the other four functions.

Fig.~5 depicts the optimization curves of \textit{DECC-G}, \textit{CMA-ES-ERDG}, \textit{LM-CMA}, \textit{CSO}, \textit{OpenAI-ES}, and our proposed \textit{DE+MF}. As can be seen, \textit{DECC-G}, \textit{CMA-ES-ERDG},  and \textit{CSO} obtained a deteriorated fitness performance among all of the methods throughout the optimization process. \textit{OpenAI-ES} and \textit{LM-CMA} achieved more promising search efficiency at the beginning of the optimization process. However, they were found to get trapped in local optima more often than the proposed method. Overall, \textit{DE+MF} obtained better fitness performance as the search progresses. These results clearly show the effectiveness and efficiency of the proposed multiform optimization method for solving the given high-dimensional problems characterized by a lower effective dimensionality.  

\begin{figure*}[!htb]
	\begin{center}
		\subfigure[Ackley Function]{\includegraphics[trim = 25mm 90mm 30mm 90mm,  clip,width=0.3\linewidth ]{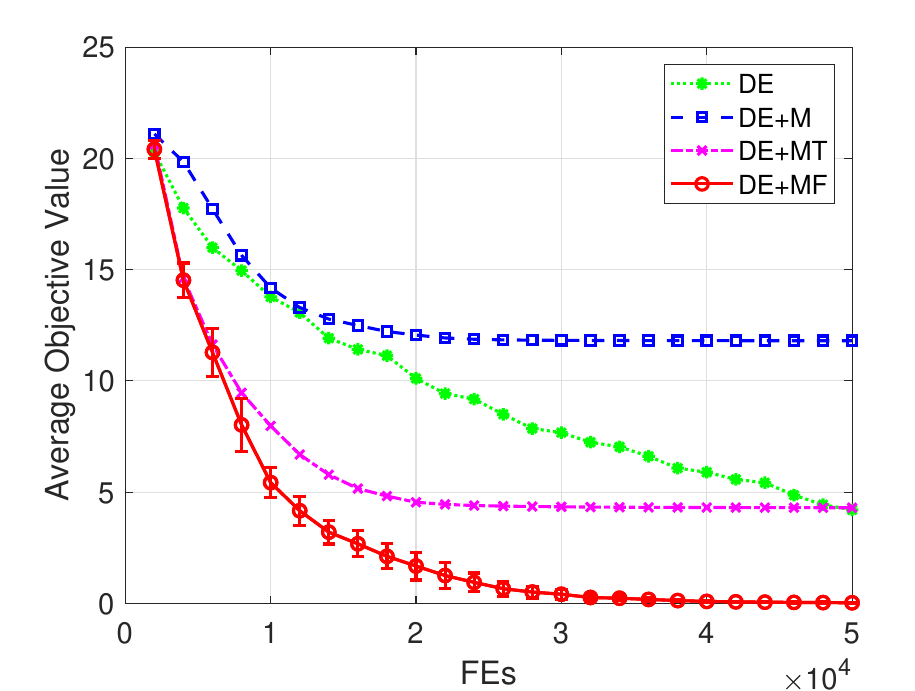}}
		\subfigure[Rastrigin Function]{\includegraphics[trim = 25mm 90mm 30mm 90mm,  clip,width=0.3\linewidth ]{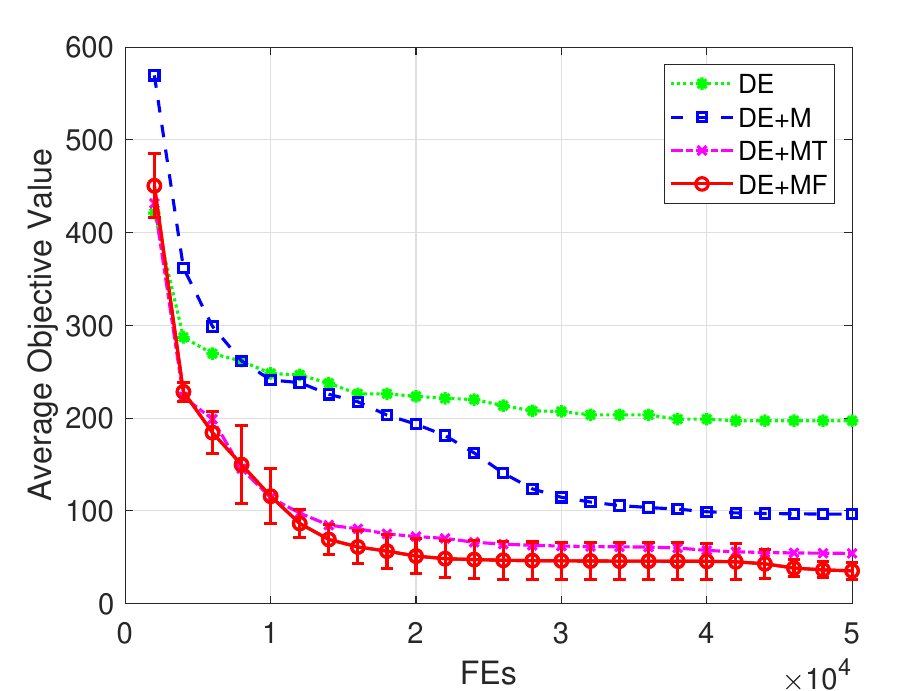}}
		\subfigure[Weierstrass Function]{\includegraphics[trim = 25mm 90mm 30mm 90mm,  clip,width=0.3\linewidth ]{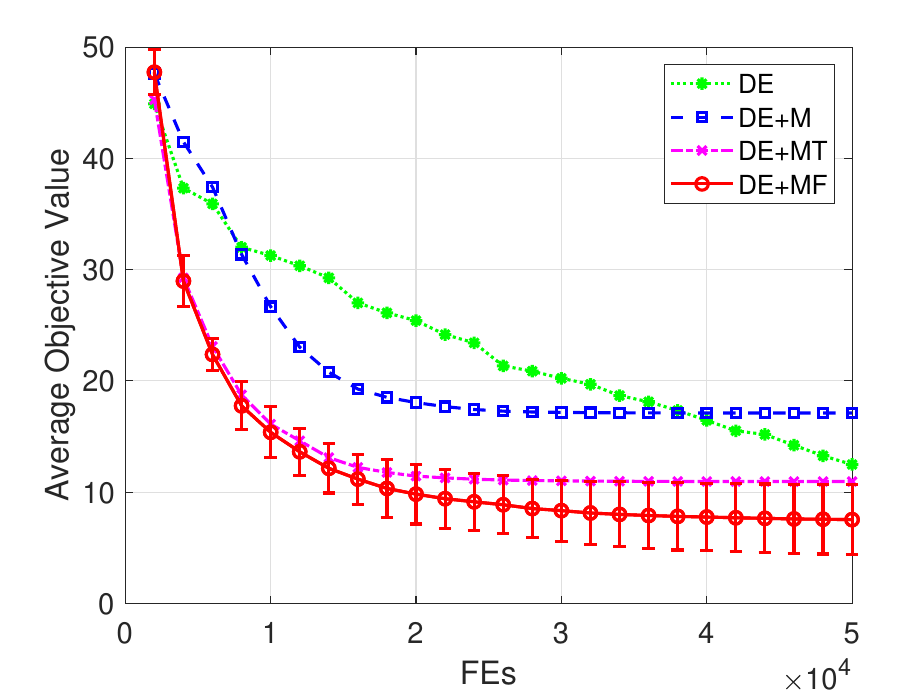}}
		\subfigure[Rosenbrock Function]{\includegraphics[trim = 25mm 90mm 30mm 90mm,  clip,width=0.3\linewidth ]{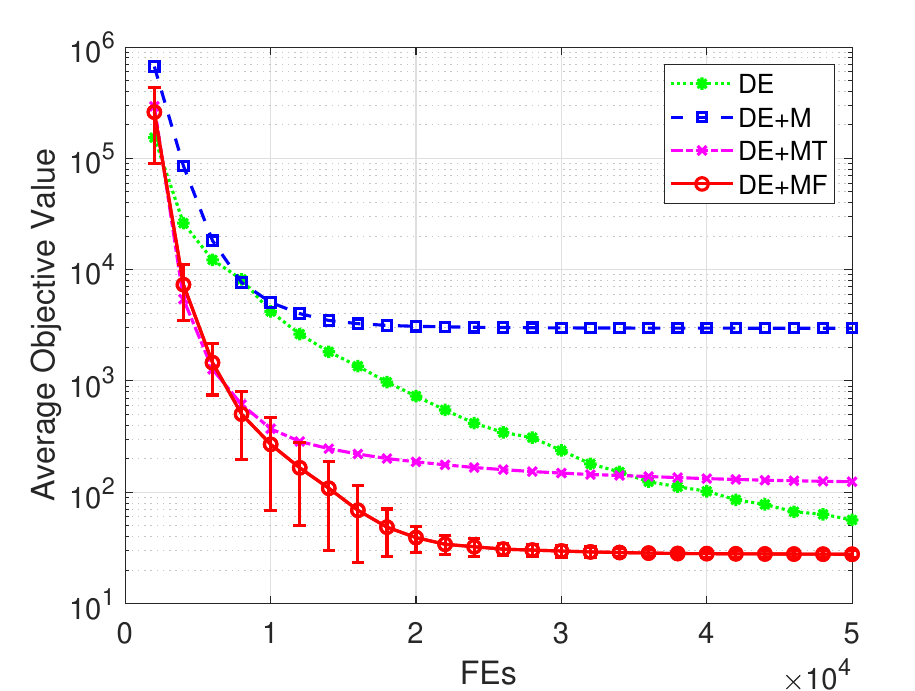}}
		\subfigure[Griewank Function]{\includegraphics[trim = 25mm 90mm 30mm 90mm,  clip,width=0.3\linewidth ]{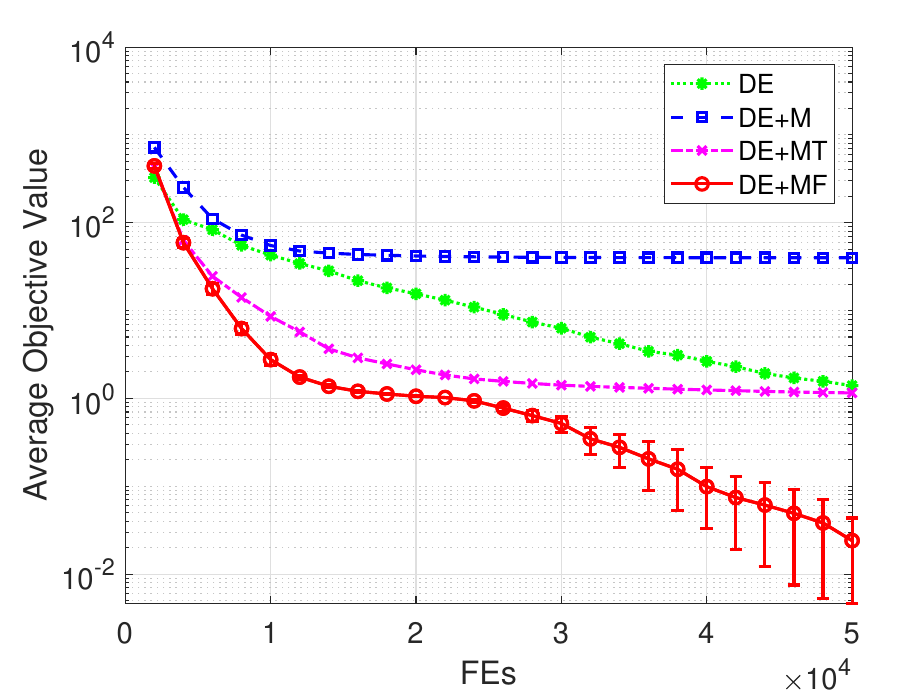}}
		\subfigure[Elliptic Function]{\includegraphics[trim = 25mm 90mm 30mm 90mm,  clip,width=0.3\linewidth ]{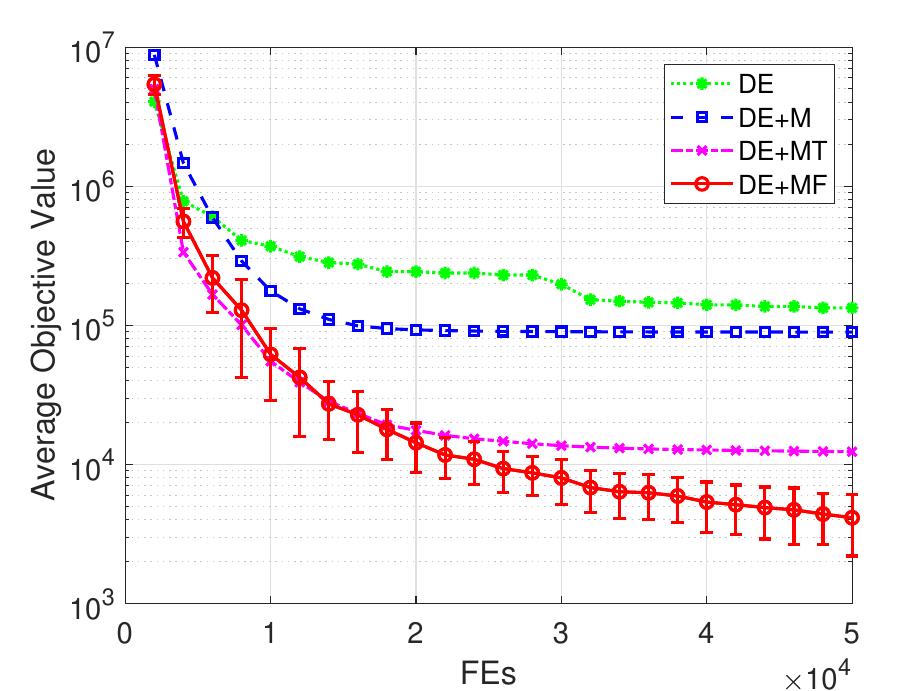}}
		\caption{Objective values obtained by \textit{DE+MF} and its variants on solving the optimization functions with $D=5000$ averaged by 20 independent runs.}
		\label{fig:App_multiformdiffD_DE}
	\end{center}
\end{figure*}

Furthermore, note that \textit{cross-form genetic transfer} and \textit{dynamic resource allocation} are two important concepts pertaining to our proposed evolutionary multiform optimization scheme. In what follows, we further verify their effectiveness in the multiform EA in an ablative manner. Specifically, the complete results of the baseline evolutionary solver (i.e.,  \textit{DE}), evolutionary solvers with low-dimensional formulations generated via random embeddings (i.e., \textit{DE+M} without cross-form genetic transfer and dynamic resource allocation), multiform evolutionary solvers with cross-form genetic transfer (i.e., \textit{DE+MT} wherein no dynamic resource allocation strategy is involved), and proposed multiform evolutionary solver (i.e., \textit{DE+MF} with both cross-form genetic transfer and dynamic resource allocation) in terms of objective values after $50,000$ function evaluations across $20$ independent runs are summarized in Table~II. For a clearer understanding, the convergence trends of the multiform EA (i.e., \textit{DE+MF}) and all of its variants are also depicted in Fig.~\ref{fig:App_multiformdiffD_DE}.

As can be observed, \textit{DE+M} obtained deteriorated performance in terms of objective values when compared to the baseline \textit{DE}. 
For example, after $50,000$ FEs, \textit{DE} reported clearly better objective values on most test functions compared to \textit{DE+M}. 
According to our definition in Section~\ref{sec:exp_A}, \textit{DE+M} optimizes the multiple low-dimensional objective formulations generated via random embeddings, and the original high-dimensional function is not embraced.  Therefore, \textit{DE+M} tends to suffer from the random embedding gaps that the global optimum of the original functions are not involved in the low-dimensional search space $\mathcal{Y}$, and hence, are trapped at the local optima (as aforementioned in Section II-C). 

Moreover, \textit{DE+MT} reported significantly faster convergence speeds during the early search stage (as shown in Fig.~\ref{fig:App_multiformdiffD_DE}) and ultimately obtained comparatively better fitness performance over the basic DE.
These superior performance highlights the efficacy of the \textit{cross-form genetic transfer} in multiform EAs by leveraging on the unique benefits (i.e., faster convergence or less local optima) offered by the other alternative formulations. Moreover, when compared to \textit{DE+MT}, the proposed \textit{DE+MF} obtained significantly better fitness performance by reporting statistically better objective values on five out of all six optimization problems. The effective and dynamic utilization of the limited computing resources can balance the optimization budget of each formulation on the fly. The formulations with a more significant convergence momentum tend to be attributed to more computing resources, and hence, provide guidance for enhancing the overall search efficiency. In summary, these results clearly verified the superiority of both\textit{ cross-form genetic transfer} and \textit{dynamic resource allocation} in our proposed multiform optimization method for solving the synthetic continuous optimization functions. 

\subsection{Additional Experimental Analysis}
In this section, more experiments are conducted to validate the effectiveness and generality of the proposed algorithm. More detailed experimental results and analysis are presented in Supplementary Materials.

First, we investigate the impact on the number of formulations in the proposed multiform optimization method. In particular, to facilitate this study, we assume the upper bound $d = 50$ of the effective dimensionality $d_e = 30$, and thereafter define the number of separate low-dimensional formulations as $N = \{1,2,3,5,10,25\}$. The experimental results show that (see Fig. 1 in Supplementary Materials), the settings on the number of low-dimensional formulations in multiform EA make a great difference in the solution quality. Fewer formulations will cause the algorithm to have a higher probability of suffering from random embedding gaps. However, when formulation numbers become large, the optimization budget (or population size) for each formulation is limited, thus leading to poor optimization performance.

Next, we take an additional step to mitigate the assumption on formulation dimensionality $d$, by considering a more realistic scenario in which multiple low-dimensional formulations bear differing dimensionality. In particular, the formulation number is set as $5$ to remain consistent with our previous study. Further, we consider the scenario in which the original optimization function and its low-dimensional formulations of $d = \{5, 25, 100, 500\}$ are involved. The corresponding experimental results show that (see Fig. 2 in Supplementary Materials), the well-evolved genetic materials transferred from formulations with a smaller search space are beneficial for complementing the search capability of formulations with more complex search spaces, thus significantly enhancing the efficiency of the overall search process.

		

\section{Practical Applications}\label{sec:applications}
In this section, we consider two practical applications of the proposed multiform optimization scheme.
Due to page limits, more detailed problem descriptions and test results are presented in Section II and III of  Supplementary Materials.

\subsection{Hyper-Parameter Tuning in Multiclass SVMs}


In machine learning, SVMs are known as powerful supervised learning models for classification challenges. When solving multi-class classification problems, one common approach is to apply a one-versus-one (OvO) strategy. For a $H$-way multiclass problem, there are $\frac{H(H-1)}{2}$ hyper-parameters to tune or optimize in all classifiers. Therefore, the dimensionality of this hyper-parameter optimization problem is $\frac{H(H-1)}{2}$. Notably, as some classes may share the same hyper-parameters, this optimization problem has a low effective dimensionality.

Hereafter, we apply the proposed multiform EA to solve the hyper-parameter tuning problem in multi-class SVMs.  In the present study, we facilitate a comparison of  \textit{DE+MF} with 1) \textit{DE} and \textit{GA}, 2) a sequential random embedding method (labeled as \textit{DE+SRE}), 3) a bayesian optimization method (\textit{REMBO}) \cite{wang2016bayesian}, 4) random search, 5) grid search, which are commonly used to optimize the hyper-parameters in SVMs. As can be observed from Table I in Supplementary Materials, both \textit{DE+SRE}, \textit{REMBO}, and \textit{DE+MF} outperform random search and grid search. This result highlights the efficacy of random embedding for tuning the hyper-parameters of multi-class SVMs in the embedded lower-dimensional search space. However, \textit{DE+MF} obtains higher testing accuracy than \textit{REMBO} and \textit{SRE} solvers. This shows that the automatic and effective knowledge transfer among multiple alternate formulations and the target problem enhances the efficiency and effectiveness of considering multiple random embeddings.

\subsection{Behavioral Predictive Model Selection in Predator-Prey Game}


In a predator-prey game, a prey agent often has to develop an efficient strategy to counter their opponent predators\cite{he2016opponent}. A feasible strategy is that the prey agent builds numerous candidate models to reason about the behavior distribution of its opponents, and identifies the most probable one to predict opponents' behaviors. This model selection process is computationally intractable as the search space of all potential models $\mathcal{M}$ could be very large~(theoretically countably infinite). Therefore, it is necessary to choose a small yet representative set of models $\mathcal{M}^{\mathcal{S}}$ from the entire model set $\mathcal{M}$ to reduce the complexity of behavior prediction. This process can be formalized as a combination optimization problem and solved with evolutionary solvers.

We take the joint behavior coverage ratio of the selected models over the full candidate model space as the evaluation index. \textcolor{black}{The performance of \textit{DE+MF} compared with 1) \textit{DE}, 2) \textit{DECC-G}), 3) \textit{GA}, 4) \textit{CMA-ES}, 5) random search, and 6) the Sub-modular Optimization (\textit{SMO}) method \cite{8944277}\cite{9712301} is investigated in two large search spaces $|\mathcal{M}|=1000, 5000$ where each candidate model is constructed as a deep neural network of different hyper-parameters.}
Compared to other methods (see Table II in Supplementary Materials), \textit{DE+MF} reported significantly higher coverage rates on selecting representative models given a certain model-size budget from 1000 and 5000 models, respectively. The performance thus highlighted the efficacy of the proposed multiform EA in solving the model selection problems with high dimensions by leveraging its multiple alternative low-dimensional formulations.

\section{Concluding Remarks and Future Work}\label{sec:con}
\textcolor{black}{In this paper, we propose an evolutionary multiform optimization method for addressing the challenges when scaling conventional EAs to high-dimensional optimization problems.} Specifically, in the present study, a realization of the evolutionary multiform optimization scheme was provided wherein multiple low-dimensional alternative formulations of a target high-dimensional problem (with low effective dimensionality) are generated via random embeddings. 

The proposed multiform EA combines all of the generated formulations into an all-encompassing multitasking environment. Most importantly, a detailed realization of a cross-from genetic transfer mechanism is presented, one that constructs connections between formulations allowing latent complementarities between them to be captured and exploited. We investigated the performance efficacy of the multiform EA via comprehensive empirical studies on both well-known continuous optimization functions as well as realistic applications. The superior performance achieved confirms the efficacy of the method in leveraging the multiple alternative formulations to significantly speed up the overall search performance in high dimensions.

Given the conceptual groundwork laid in this paper, several opportunities emerge for future research development. The general theme of our approach has been to piggyback on simple tasks to solve related, but significantly more complex, problems.  This basic idea naturally gives rise to many other practical applications of the multiform paradigm - not limited to dimensionality reductions. In addition, extensions to the case of multiple and many-objective problems will be considered in the near future.

\bibliographystyle{IEEEtran}
\bibliography{IEEEabrv,./References}

\begin{thebibliography}{10}
\providecommand{\url}[1]{#1}
\csname url@samestyle\endcsname
\providecommand{\newblock}{\relax}
\providecommand{\bibinfo}[2]{#2}
\providecommand{\BIBentrySTDinterwordspacing}{\spaceskip=0pt\relax}
\providecommand{\BIBentryALTinterwordstretchfactor}{4}
\providecommand{\BIBentryALTinterwordspacing}{\spaceskip=\fontdimen2\font plus
\BIBentryALTinterwordstretchfactor\fontdimen3\font minus \fontdimen4\font\relax}
\providecommand{\BIBforeignlanguage}[2]{{%
\expandafter\ifx\csname l@#1\endcsname\relax
\typeout{** WARNING: IEEEtran.bst: No hyphenation pattern has been}%
\typeout{** loaded for the language `#1'. Using the pattern for}%
\typeout{** the default language instead.}%
\else
\language=\csname l@#1\endcsname
\fi
#2}}
\providecommand{\BIBdecl}{\relax}
\BIBdecl

\bibitem{back1996}
T.~Back, \emph{Evolutionary algorithms in theory and practice: evolution strategies, evolutionary programming, genetic algorithms}.\hskip 1em plus 0.5em minus 0.4em\relax Oxford university press, 1996.

\bibitem{8356195}
W.~Gong, Y.~Wang, Z.~Cai, and L.~Wang, ``Finding multiple roots of nonlinear equation systems via a repulsion-based adaptive differential evolution,'' \emph{IEEE Transactions on Systems, Man, and Cybernetics: Systems}, vol.~50, no.~4, pp. 1499--1513, 2020.

\bibitem{8456559}
Y.~Jin, H.~Wang, T.~Chugh, D.~Guo, and K.~Miettinen, ``Data-driven evolutionary optimization: An overview and case studies,'' \emph{IEEE Transactions on Evolutionary Computation}, vol.~23, no.~3, pp. 442--458, 2019.

\bibitem{zhan2022survey}
Z.-H. Zhan, L.~Shi, K.~C. Tan, and J.~Zhang, ``A survey on evolutionary computation for complex continuous optimization,'' \emph{Artificial Intelligence Review}, vol.~55, no.~1, pp. 59--110, 2022.

\bibitem{ma2018survey}
X.~Ma, X.~Li, Q.~Zhang, K.~Tang, Z.~Liang, W.~Xie, and Z.~Zhu, ``A survey on cooperative co-evolutionary algorithms,'' \emph{IEEE Transactions on Evolutionary Computation}, vol.~23, no.~3, pp. 421--441, 2018.

\bibitem{liu2021comprehensive}
S.~Liu, Q.~Lin, Q.~Li, and K.~C. Tan, ``A comprehensive competitive swarm optimizer for large-scale multiobjective optimization,'' \emph{IEEE Transactions on Systems, Man, and Cybernetics: Systems}, 2021.

\bibitem{deng2021quantum}
W.~Deng, S.~Shang, X.~Cai, H.~Zhao, Y.~Zhou, H.~Chen, and W.~Deng, ``Quantum differential evolution with cooperative coevolution framework and hybrid mutation strategy for large scale optimization,'' \emph{Knowledge-Based Systems}, vol. 224, p. 107080, 2021.

\bibitem{Lever2017}
J.~Lever, M.~Krzywinski, and N.~Altman, ``Principal component analysis,'' \emph{Nature Methods}, vol.~14, no.~7, pp. 641--642, Jul 2017.

\bibitem{10.1007/BFb0020217}
B.~Sch{\"o}lkopf, A.~Smola, and K.-R. M{\"u}ller, ``Kernel principal component analysis,'' in \emph{Artificial Neural Networks --- ICANN'97}, W.~Gerstner, A.~Germond, M.~Hasler, and J.-D. Nicoud, Eds.\hskip 1em plus 0.5em minus 0.4em\relax Berlin, Heidelberg: Springer Berlin Heidelberg, 1997, pp. 583--588.

\bibitem{10.5555/944919.944937}
D.~M. Blei, A.~Y. Ng, and M.~I. Jordan, ``Latent dirichlet allocation,'' \emph{J. Mach. Learn. Res.}, vol.~3, no. null, p. 993–1022, mar 2003.

\bibitem{9627116}
M.~N. Omidvar, X.~Li, and X.~Yao, ``A review of population-based metaheuristics for large-scale black-box global optimization—part i,'' \emph{IEEE Transactions on Evolutionary Computation}, vol.~26, no.~5, pp. 802--822, 2022.

\bibitem{li2018cooperative}
K.~Li, Q.~Liu, S.~Yang, J.~Cao, and G.~Lu, ``Cooperative optimization of dual multiagent system for optimal resource allocation,'' \emph{IEEE Transactions on systems, man, and cybernetics: systems}, vol.~50, no.~11, pp. 4676--4687, 2018.

\bibitem{sun2019hybrid}
L.~Sun, L.~Lin, M.~Gen, and H.~Li, ``A hybrid cooperative coevolution algorithm for fuzzy flexible job shop scheduling,'' \emph{IEEE Transactions on Fuzzy Systems}, vol.~27, no.~5, pp. 1008--1022, 2019.

\bibitem{shi2021coevolutionary}
W.~Shi, W.-N. Chen, S.~Kwong, J.~Zhang, H.~Wang, T.~Gu, H.~Yuan, and J.~Zhang, ``A coevolutionary estimation of distribution algorithm for group insurance portfolio,'' \emph{IEEE Transactions on Systems, Man, and Cybernetics: Systems}, 2021.

\bibitem{2016A}
Y.~Mei, M.~N. Omidvar, X.~Li, and X.~Yao, ``A competitive divide-and-conquer algorithm for unconstrained large-scale black-box optimization,'' \emph{ACM Transactions on Mathematical Software (TOMS)}, 2016.

\bibitem{7911173}
M.~N. Omidvar, M.~Yang, Y.~Mei, X.~Li, and X.~Yao, ``Dg2: A faster and more accurate differential grouping for large-scale black-box optimization,'' \emph{IEEE Transactions on Evolutionary Computation}, vol.~21, no.~6, pp. 929--942, 2017.

\bibitem{9141328}
M.~Yang, A.~Zhou, C.~Li, and X.~Yao, ``An efficient recursive differential grouping for large-scale continuous problems,'' \emph{IEEE Transactions on Evolutionary Computation}, vol.~25, no.~1, pp. 159--171, 2021.

\bibitem{rusu2018meta}
A.~A. Rusu, D.~Rao, J.~Sygnowski, O.~Vinyals, R.~Pascanu, S.~Osindero, and R.~Hadsell, ``Meta-learning with latent embedding optimization,'' \emph{arXiv preprint arXiv:1807.05960}, 2018.

\bibitem{kaban2016toward}
A.~Kab{\'a}n, J.~Bootkrajang, and R.~J. Durrant, ``Toward large-scale continuous eda: A random matrix theory perspective,'' \emph{Evolutionary computation}, vol.~24, no.~2, pp. 255--291, 2016.

\bibitem{wang2016bayesian}
Z.~Wang, F.~Hutter, M.~Zoghi, D.~Matheson, and N.~de~Feitas, ``Bayesian optimization in a billion dimensions via random embeddings,'' \emph{Journal of Artificial Intelligence Research}, vol.~55, pp. 361--387, 2016.

\bibitem{qian2016derivative}
H.~Qian, Y.-Q. Hu, and Y.~Yu, ``Derivative-free optimization of high-dimensional non-convex functions by sequential random embeddings.'' in \emph{IJCAI}, 2016, pp. 1946--1952.

\bibitem{yang2021parallel}
Q.~Yang, P.~Yang, and K.~Tang, ``Parallel random embedding with negatively correlated search,'' in \emph{International Conference on Swarm Intelligence}.\hskip 1em plus 0.5em minus 0.4em\relax Springer, 2021, pp. 339--351.

\bibitem{gupta2016multifactorial}
A.~Gupta, Y.-S. Ong, and L.~Feng, ``Multifactorial evolution: toward evolutionary multitasking,'' \emph{IEEE Transactions on Evolutionary Computation}, vol.~20, no.~3, pp. 343--357, 2016.

\bibitem{tan2021evolutionary}
K.~C. Tan, L.~Feng, and M.~Jiang, ``Evolutionary transfer optimization-a new frontier in evolutionary computation research,'' \emph{IEEE Computational Intelligence Magazine}, vol.~16, no.~1, pp. 22--33, 2021.

\bibitem{gupta2018insights}
A.~Gupta, Y.-S. Ong, and L.~Feng, ``Insights on transfer optimization: Because experience is the best teacher,'' \emph{IEEE Transactions on Emerging Topics in Computational Intelligence}, vol.~2, no.~1, pp. 51--64, 2018.

\bibitem{qian2016scaling}
H.~Qian and Y.~Yu, ``Scaling simultaneous optimistic optimization for high-dimensional non-convex functions with low effective dimensions.'' in \emph{AAAI}, 2016, pp. 2000--2006.

\bibitem{tu2019autone}
K.~Tu, J.~Ma, P.~Cui, J.~Pei, and W.~Zhu, ``Autone: Hyperparameter optimization for massive network embedding,'' in \emph{Proceedings of the 25th ACM SIGKDD International Conference on Knowledge Discovery \& Data Mining}, 2019, pp. 216--225.

\bibitem{narayanan2021randomized}
S.~Narayanan, S.~Silwal, P.~Indyk, and O.~Zamir, ``Randomized dimensionality reduction for facility location and single-linkage clustering,'' in \emph{International Conference on Machine Learning}.\hskip 1em plus 0.5em minus 0.4em\relax PMLR, 2021, pp. 7948--7957.

\bibitem{8100935}
M.~Jiang, Z.~Huang, L.~Qiu, W.~Huang, and G.~G. Yen, ``Transfer learning-based dynamic multiobjective optimization algorithms,'' \emph{IEEE Transactions on Evolutionary Computation}, vol.~22, no.~4, pp. 501--514, 2018.

\bibitem{tian2019bi}
X.~Tian and H.~Niu, ``A bi-objective model with sequential search algorithm for optimizing network-wide train timetables,'' \emph{Computers \& Industrial Engineering}, vol. 127, pp. 1259--1272, 2019.

\bibitem{xue2020affine}
X.~Xue, K.~Zhang, K.~C. Tan, L.~Feng, J.~Wang, G.~Chen, X.~Zhao, L.~Zhang, and J.~Yao, ``Affine transformation-enhanced multifactorial optimization for heterogeneous problems,'' \emph{IEEE Transactions on Cybernetics}, 2020.

\bibitem{zhong2018multifactorial}
J.~Zhong, L.~Feng, W.~Cai, and Y.-S. Ong, ``Multifactorial genetic programming for symbolic regression problems,'' \emph{IEEE transactions on systems, man, and cybernetics: systems}, vol.~50, no.~11, pp. 4492--4505, 2018.

\bibitem{liang2021multiobjective}
Z.~Liang, W.~Liang, Z.~Wang, X.~Ma, L.~Liu, and Z.~Zhu, ``Multiobjective evolutionary multitasking with two-stage adaptive knowledge transfer based on population distribution,'' \emph{IEEE Transactions on Systems, Man, and Cybernetics: Systems}, 2021.

\bibitem{feng2021multi}
L.~Feng, Q.~Shang, Y.~Hou, K.~C. Tan, and Y.-S. Ong, ``Multi-space evolutionary search for large-scale optimization,'' \emph{arXiv preprint arXiv:2102.11693}, 2021.

\bibitem{chen2020evolutionary}
K.~Chen, B.~Xue, M.~Zhang, and F.~Zhou, ``An evolutionary multitasking-based feature selection method for high-dimensional classification,'' \emph{IEEE Transactions on Cybernetics}, 2020.

\bibitem{zhang2021study}
L.~Zhang, Y.~Xie, J.~Chen, L.~Feng, C.~Chen, and K.~Liu, ``A study on multiform multi-objective evolutionary optimization,'' \emph{Memetic Computing}, pp. 1--12, 2021.

\bibitem{feng2017autoencoding}
L.~Feng, Y.-S. Ong, S.~Jiang, and A.~Gupta, ``Autoencoding evolutionary search with learning across heterogeneous problems,'' \emph{IEEE Transactions on Evolutionary Computation}, vol.~21, no.~5, pp. 760--772, 2017.

\bibitem{8401802}
L.~Feng, L.~Zhou, J.~Zhong, A.~Gupta, Y.-S. Ong, K.-C. Tan, and A.~K. Qin, ``Evolutionary multitasking via explicit autoencoding,'' \emph{IEEE Transactions on Cybernetics}, vol.~49, no.~9, pp. 3457--3470, 2019.

\bibitem{feng2020large}
Y.~Feng, L.~Feng, Y.~Hou, and K.~C. Tan, ``Large-scale optimization via evolutionary multitasking assisted random embedding,'' in \emph{2020 IEEE Congress on Evolutionary Computation (CEC)}.\hskip 1em plus 0.5em minus 0.4em\relax IEEE, 2020, pp. 1--8.

\bibitem{gong2019evolutionary}
M.~Gong, Z.~Tang, H.~Li, and J.~Zhang, ``Evolutionary multitasking with dynamic resource allocating strategy,'' \emph{IEEE Transactions on Evolutionary Computation}, 2019.

\bibitem{9570733}
T.~Wei and J.~Zhong, ``Towards generalized resource allocation on evolutionary multitasking for multi-objective optimization,'' \emph{IEEE Computational Intelligence Magazine}, vol.~16, no.~4, pp. 20--37, 2021.

\bibitem{9493747}
N.~Zhang, A.~Gupta, Z.~Chen, and Y.-S. Ong, ``Evolutionary machine learning with minions: A case study in feature selection,'' \emph{IEEE Transactions on Evolutionary Computation}, vol.~26, no.~1, pp. 130--144, 2022.

\bibitem{9950429}
------, ``Multitask neuroevolution for reinforcement learning with long and short episodes,'' \emph{IEEE Transactions on Cognitive and Developmental Systems}, vol.~15, no.~3, pp. 1474--1486, 2023.

\bibitem{9943989}
Z.~Chen, A.~Gupta, L.~Zhou, and Y.-S. Ong, ``Scaling multiobjective evolution to large data with minions: A bayes-informed multitask approach,'' \emph{IEEE Transactions on Cybernetics}, pp. 1--14, 2022.

\bibitem{storn1997}
R.~Storn and K.~Price, ``Differential evolution--a simple and efficient heuristic for global optimization over continuous spaces,'' \emph{Journal of global optimization}, vol.~11, no.~4, pp. 341--359, 1997.

\bibitem{Deb1995SimulatedBC}
K.~Deb and R.~B. Agrawal, ``Simulated binary crossover for continuous search space,'' \emph{Complex Syst.}, vol.~9, 1995.

\bibitem{he2016opponent}
H.~He, J.~Boyd-Graber, K.~Kwok, and H.~Daum{\'e}~III, ``Opponent modeling in deep reinforcement learning,'' in \emph{International conference on machine learning}.\hskip 1em plus 0.5em minus 0.4em\relax PMLR, 2016, pp. 1804--1813.

\bibitem{8944277}
Y.~Hou, Y.-S. Ong, J.~Tang, and Y.~Zeng, ``Evolutionary multiagent transfer learning with model-based opponent behavior prediction,'' \emph{IEEE Transactions on Systems, Man, and Cybernetics: Systems}, vol.~51, no.~10, pp. 5962--5976, 2021.

\bibitem{9712301}
Y.~Hou, M.~Sun, W.~Zhu, Y.~Zeng, H.~Piao, X.~Chen, and Q.~Zhang, ``Behavior reasoning for opponent agents in multi-agent learning systems,'' \emph{IEEE Transactions on Emerging Topics in Computational Intelligence}, vol.~6, no.~5, pp. 1125--1136, 2022.

\end{thebibliography}

\newpage

\end{document}